\documentclass[11pt]{article}
\usepackage[utf8]{inputenc}
\usepackage[margin=1.0in]{geometry}  % margins

\usepackage{amsmath}
\usepackage{amssymb}
\usepackage{bm}

\usepackage{cite}

\usepackage[title]{appendix}

\usepackage{xcolor}
\usepackage{enumitem}

\usepackage{graphics} % for pdf, bitmapped graphics files
\usepackage{epsfig} % for postscript graphics files

\usepackage[caption=false,font=footnotesize]{subfig}%\fi
\usepackage[export]{adjustbox}
\usepackage{multicol} % added by ls

\title{\huge{Fusion of Heterogeneous Friction Estimates for \\ Traction Adaptive Motion Planning and Control}}
\author{Lars Svensson and Martin T\"orngren% <-this % stops a space
\thanks{L. Svensson and M. T\"orngren are with the Department of Machine Design, KTH Royal Institute of
Technology, Stockholm, Sweden. {\tt\small \{larsvens,martint\}@kth.se}}%
}

\date{\today}

\begin{document}
\maketitle
\begin{abstract}
% motivation and scope
Traction adaptive motion planning and control has potential to improve an an automated vehicle's ability to avoid accident in a critical situation. However, such functionality require an accurate friction estimate for the road ahead of the vehicle that is updated in real time. Current state of the art friction estimation techniques include high accuracy local friction estimation in the presence of tire slip, as well as rough classification of the road surface ahead of the vehicle, based on forward looking camera. In this paper we show that neither of these techniques in isolation yield satisfactory behavior when deployed with traction adaptive motion planning and control functionality. However, fusion of the two provides sufficient accuracy, availability and foresight to yield near optimal behavior. To this end, we propose a fusion method based on heteroscedastic gaussian process regression, and present initial simulation based results. 
\end{abstract}

\section{Introduction}
\label{sec:intro}

%Paper Story: 
% we have tamp. raises question: sensitivity to uncertainty in predictive estimate 
% multiple sensors --> GP rep --> can still make aggressive or conservative plans, but within interval that gets updated based on new data. 
% How to best integrate multiple estimates? --> GP
% concept: traction uncertainty awareness - awareness of the uncertainty of predictive friction estimate and systematic selection of conservative or aggressive policies

% RQ: what is the impact of traction awareness on accident mitigation performance in CS? 

% critical situations
One of the remaining challenges in the development of automated driving (AD) and advanced driver assistance systems (ADAS) is fully automated operation in critical situations i.e., situations where an accident is imminent, for example due to unpredictable behavior of other traffic agents or rapid changes in the operational conditions. In critical situations, full utilization of the physical capacity of the vehicle is paramount to minimizing the risk of an accident, since overly cautious behavior will impair accident mitigation performance. This motivates research in motion planning and control at the limits of handling - a field that has seen a lot of progress in the last decade \cite{berntorp2014models,liniger2015optimization,funke2017collision,subosits2019qcqp,fors2020autonomous}. 
At the limit of tire adhesion, longitudinal, lateral and yaw dynamics are tightly coupled. Therefore, the traditional approach of performing motion planning, lateral- and longitudinal control in isolation is being outperformed by combined optimization approaches, where the lateral and longitudinal motion planning and control is performed by solving a single optimization problem \cite{liniger2015optimization,subosits2019qcqp}. 

% traction adapt - ref tcst paper 
To further complicate the problem of motion planning and control in critical situations, the physical motion capacity of a vehicle is far from static. Instead, it varies greatly with the operational situation, e.g. road surface and tire conditions. In our previous work, we tackle the problem of motion planning and control under time-varying traction constraints \cite{svensson2019adaptive,svensson2020traction}. 
% requires friction estimation and acting under uncertainty! 
We show that such an approach improves the vehicle's ability to avoid accident in critical situations, and thus have potential to improve safety performance of AD and ADAS systems.
However, the performance increase hinges on run-time availability of a tire-road friction estimate over the planning horizon, i.e., \textit{ahead} of the vehicle. We refer to this concept as a \textit{predictive friction estimate}. 
% Highlight the problem of conservative but not overly conservative tire force constraints! 
One of the conclusions drawn from our previous work is that in critical situations, accident avoidance performance is negatively affected by both over- and under-estimation in the predictive friction estimate. Over-estimation leads to the planning of dynamically infeasible motions, and under-estimation to poor utilization of the available traction, which can needlessly lead to accidents in e.g., a collision avoidance situation. To summarize, accurate predictive friction estimation is key for autonomous accident mitigation at the limits of handling. 

\begin{figure}[t]
    \centering
    \includegraphics[width=0.75\textwidth]{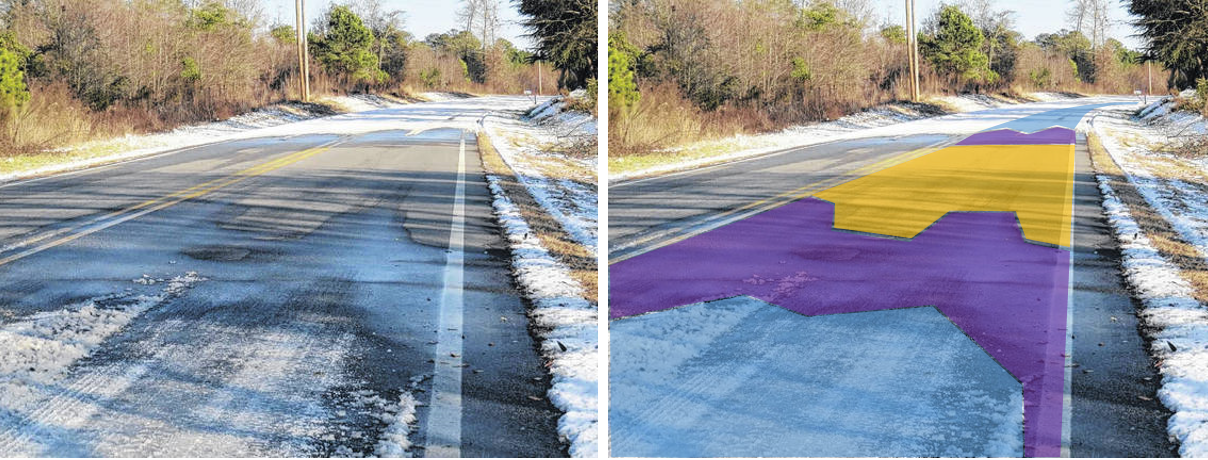}
    \caption{Illustration road surface classification within a lane, based on forward looking camera. Classes: orange - dry, purple - wet, blue - snow/ice}
    \label{fig:cam_fig}
\end{figure}

% Highlight the problems of current local and predictive traction estimation paradigms
% (uncertainty, availability)

Fortunately, research over the past two decades have presented several viable methods of obtaining such estimates \cite{khaleghian2017technical,andersson2007road}. In later years, the rapid development in supervised machine learning and computer vision have greatly improved performance of such functionality, mainly through road surface classification from forward looking camera images \cite{sohini2018predfrictionest,nolte2018predfrictionest}, providing a predictive estimate online with high availability. However, due to the limited number of visually distinguishable road surface classes (e.g., dry, wet, snow/ice), accuracy is  poor compared to state of the art local friction estimation methods. 
% local
Local friction estimation methods \cite{rajamani2012algorithms,gustafsson1997slip} use vehicle dynamics models and traditional estimation techniques to estimate the friction coefficient locally around the vehicle. Such methods have high accuracy but the estimate is only available in the presence of tire slip \cite{prokevs2016quantification}, and provide no foresight about the conditions ahead of the vehicle. 
% Explain why this is a problem wrt tamp
Considering the high level properties of the two paradigms;
\begin{itemize}
    \item Camera based: high availability, foresight, low accuracy,
    \item Local: low availability, lack of foresight, high accuracy,
\end{itemize}
we hypothesize that using either method individually could lead to poor performance in the context of traction adaptive motion planning. However, the indication that the two paradigms are to some extent complementary, motivates investigation in fusion of the two.  

%We hypothesise that in the context of motion planning and control at the limits of handling, awareness of the certainty level in the friction estimate can improve performance in critical situations. 

% RELATED WORK (GPs in MP)
Gaussian process (GP) regression \cite{rasmussen2003gaussian} is a nonparametric Bayesian approach to regression, that has become an increasingly popular modeling tool in optimization based control. In contrast to the traditional approach of modelling the system to be controlled from first principles, a GP representation of the system does not make strong assumptions of the underlying dynamics, instead the dynamics, including a direct representation of model uncertainty, is inferred from input/output data. Previous approaches uses GP representations either to completely replace, \cite{kocijan2003predictive,cao2017gaussian,kamthe2018data,rodriguez2021learning} or to complement a traditional dynamics model  \cite{hewing2019cautious,kabzan2019learning,mckinnon2018experience,ostafew2016robust}, by representing residual dynamics and uncertainty.

% --> How ours differ, why this is good
This work differentiates from the previous efforts by using a GP representation of time varying input constraints, rather than the dynamics. We extend the framework of \cite{svensson2020traction}, with a novel, GP regression based method for merging predictive and local friction estimates, and quantitatively represent uncertainty in the fused estimate. We show that such a fusion strategy is able to exploit the virtues of both camera based and local friction estimation. 
In summary, the contributions of this paper are: 
\begin{itemize}
\item Justification for the argument that neither camera based nor local state of the art friction estimation techniques in isolation is sufficient for traction adaptive motion planning and control.
\item Preliminary results indicating that fusion of the two complementary estimation paradigms can produce near-optimal behavior.
\item A proposed GP regression based method for fusion of predictive and local friction estimates that benefits from the virtues of both paradigms. 
\end{itemize}

\section{Background}
\label{sec:tamp_summary}
Here follows a summary of the traction adaptive motion planning and control framework which forms a basis of this work. See \cite{svensson2020traction} for a detailed account.
At each planning time $t$, a constrained finite time optimal control problem 
\begin{equation*}
\begin{array}{ll}
\underset{u_{0|t}, \dots, u_{N-1|t}}{\mbox{min}} & J(\mathcal{T}_t)   \\\
~~~~~~\mbox{s.t.,} 	& x_{k+1|t} = f \left(x_{k|t},u_{k|t} \right), \\
			    & u_{k|t} \in \mathcal{U}_{k|t}, \\
			    & x_{k|t} \in \mathcal{X}_{k|t}, \\ 
			    & \forall k \in \{0,1,\dots,(N-1)\},\\
			    & x_{0|t} = x_t,~x_{N|t} \in \mathcal{X}_{k|t},  
\end{array}
\label{eq:cftoc}
\end{equation*}
\sloppy is solved over a prediction horizon $k \in \{0,1,\dots, (N-1)\}$. The predicted input sequence  $\{u_{0|t},u_{1|t}, \dots,u_{N-1|t}\}$ is applied to the vehicle model $f(\cdot,\cdot)$, yielding the corresponding predicted state sequence $\{x_{0|t},x_{1|t}, \dots, x_{N|t}\}$.
We compactly denote a predicted trajectory consisting of state and control sequences as $\mathcal{T}_t = \{\{x_{k|t}\}_{k=0}^{N}, \{ u_{k|t}\}_{k=0}^{N-1}\}$. 
The vehicle model $f(\cdot,\cdot)$ is a dynamic bicycle model with tire force inputs, expressed in a road aligned coordinate frame (See Appendix B of \cite{svensson2020traction}). 
The behavior of the resulting optimal trajectory $\mathcal{T}^\star_t$ is dictated by a cost function $J(\mathcal{T}_t)$ and time varying state and input constraints $\mathcal{X}_{k|t}$, $\mathcal{U}_{k|t}$. State constraints $\mathcal{X}_{k|t}$ are used to encode lane boundaries as well as static and dynamic obstacles. 

We use the time varying input constraint $\mathcal{U}_{k|t}$ as a mechanism to encode the tire adhesion limit in the presence of local traction variations. 
For each timestep $k$ in the prediction horizon, tire force inputs are upper bounded by a convex polygon. The size and shape of each polygon is set based on a tire model, predicted normal forces and a predictive friction estimate $\hat{\mu}(s)$. 

% For each timestep $k$, the upper bound of the front and back tire forces are computed as 
% \begin{align*}
%     F_\mathrm{f}^{(\mathrm{ub})} = \lambda \mu_{\text{est}} F_{z\mathrm{f}}, \quad
%     F_\mathrm{r}^{(\mathrm{ub})} = \lambda \mu_{\text{est}} F_{z\mathrm{r}},
% \end{align*}
% where $\lambda$ is a tunable force utilization factor, $\mu_{\text{est}}$ is the predictive friction estimate at $s_k$ and $F_{z\mathrm{(f/r)}}$ are the predicted normal forces at $k$.

% Description of the method
\section{Fusing Heterogeneous Friction \\ Estimates through GP Regression}
\label{sec:method}
\begin{figure}[t]
    \centering
    \includegraphics[width=0.6\textwidth]{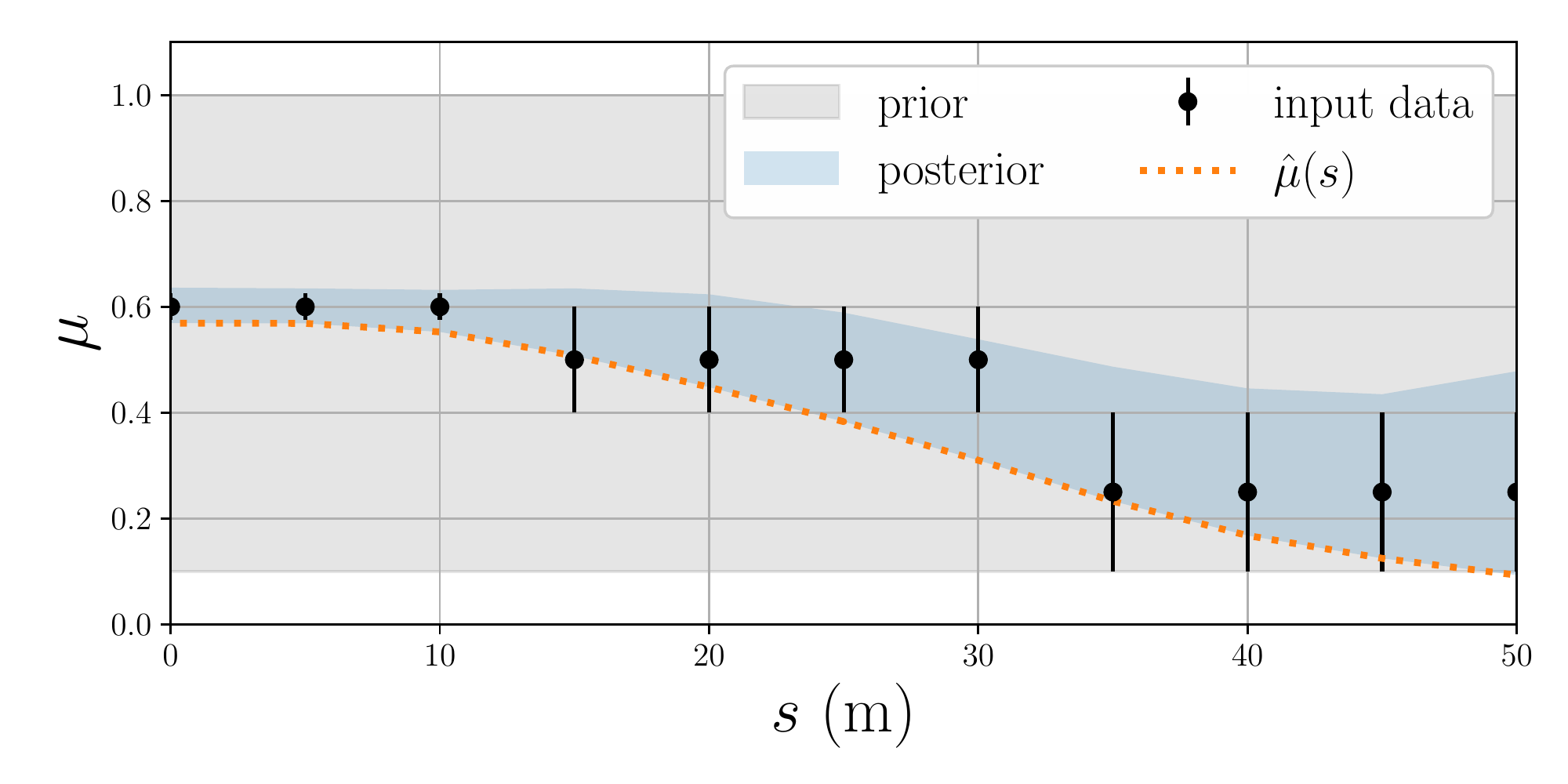}
    \caption{Example showing how the proposed method merges domain knowledge encoded in the prior (gray) with local and predictive friction estimates (black dots) of varying uncertainty (black bars). The resulting estimate is selected as the lower bound of the 95\% confidence region of the posterior distribution}
    \label{fig:fusion_full}
\end{figure}

% aim of the thing
The aim of the proposed fusion method is to obtain a vector $\hat{\mu}(s)$, that represents the tire-road friction at a sequence of discrete distances ahead of the vehicle, $s \in \{0,ds,\dots, s_f\}$, with $s_f$ corresponding to the spatial extension of the prediction horizon of the planning/control framework introduced in Section \ref{sec:tamp_summary}. The estimate should be
\begin{enumerate}[label=(\alph*)]
    \item Conservative - in the sense that the probability of over-estimation is low. This is a generally applied principle in using online friction estimates for vehicle control, because over-estimation of the friction coefficient may lead to loss of control of the vehicle \cite{rajamani2011vehicle,yi2002underest}.
    \item Not overly conservative - to avoid that collision avoidance performance deteriorates, e.g., in the manner described in \cite{svensson2020traction}.
\end{enumerate}
% general introduction to the appraoch
Our proposed fusion strategy is to infer a distribution over functions $f_\star (s)$ at $s \in \{0,ds,\dots, s_f\}$, that with high probability encompasses the true predictive friction values $\mu(s)$. The distribution is obtained through heteroscedastic GP regression based on input data from predictive \textit{and} local friction estimates, including their associated estimation uncertainties, which are propagated from the individual methods.  
From such a distribution $f_\star(s)$ we then select a $\hat{\mu}(s)$ that satisfies both (a) and (b). See Fig.~\ref{fig:fusion_full} for a visual example and Appendix \ref{sec:gp_preliminaries} for a brief outline of GP regression preliminaries.
% \begin{equation*}
%     f(s) \sim \mathcal{GP}(\eta(s),k(s,s')).
% \end{equation*}

% step 1: Assign a prior
We start by selecting a prior $f(s) \sim \mathcal{GP}(\eta(s),k(s,s'))$, with mean function $\eta(s)$ and kernel function $k(s,s')$, 
based on domain knowledge, i.e that it is probable that $\mu(s)$ resides in the interval 0.1 to 1.0, and that some spatial correlation exists, i.e., that positions close to each other are likely to have similar friction coefficients. 
For $k(s,s')$, we use a squared exponential kernel, Equation \eqref{eq:squared_exp_kernel} of Appendix \ref{sec:gp_preliminaries}, and select $\eta(s)$ and $\sigma_f$ of \eqref{eq:squared_exp_kernel} such that the interval $[0.1, 1.0]$ represents a 95\% confidence interval for the prior distribution over functions, as illustrated in Fig.~\ref{fig:fusion_full}. 
The prior is determined offline, in contrast to the subsequent steps that are to be executed online.

% step 2: set the input data 
Next, we prepare the input data. We combine friction estimates and associated margins of error from predictive and local friction estimation algorithms into two input data vectors 
\begin{equation*}
  \hat{\mu}'(s), m'(s) =
    \begin{cases}
      \hat{\mu}_l, m_l ~~~~ \text{if local available \& } s<s_{l}, \\
      \hat{\mu}_p(s), m_p(s) ~~~ \text{otherwise}.
    \end{cases}       
\end{equation*}
Subscripts $(\cdot)_l$ and $(\cdot)_p$ indicate local and predictive estimates respectively.
The predictive friction estimate $\hat{\mu}_p(s)$ is set as the mean value of its road surface class along with a margin of error $m_p(s)$ set such that for all $s \in \{0,ds,\dots, s_f\}$, $\hat{\mu}_p(s) \pm m_p(s)$ encompasses all friction coefficient values within the road surface class associated with $s$. If tire slip is sufficient to make a local estimate $\hat{\mu}_l$ available, the predictive estimates at $s \in \{0,ds,\dots, s_{l}\}$ are replaced by $\hat{\mu}_l$ and $m_l$. The local margin of error $m_l$ is set as the worst case estimation error of the local estimation method. A tunable threshold $s_{l}$ dictates how far ahead the local estimate will influence the input data.

% interpret margin of error as a 95% confidence interval --> \sigmas

% step 3: compute posterior
Next, we compute the mean function and covariance matrix of the posterior predictive distribution $f_\star(s) \sim \mathcal{GP}(\eta_\star(s),\Sigma_\star)$
at locations $s \in \{0,ds,\dots, s_f\}$, from the prior and the input data $\hat{\mu}'(s)$ and $m'(s)$ using equations \eqref{eq:gp_post_mean} and \eqref{eq:gp_post_cov} of Appendix \ref{sec:gp_preliminaries}. Here, we interpret the interval $\hat{\mu}'(s) \pm m'(s)$ as a 95\% confidence interval to obtain the standard deviation of the input data $\sigma'(s) = m'(s)/1.96$.
Finally, we select the fused predictive friction estimate as the lower bound of the 95\% confidence region of $f_\star(s)$.
\begin{align*}
    &\hat{\mu}(s) = \eta_\star(s) - 1.96\sigma_\star(s) \quad \text{with} \\
    &\sigma_\star(s) = \sqrt{\text{diag}(\Sigma_\star)}. 
\end{align*}
Fig.~\ref{fig:fusion_full} shows an example plot of the prior distribution, the input data, the posterior distribution and the final predictive friction estimate for a single planning iteration, in a scenario where the vehicle is approaching a low $\mu$ section.
The resulting estimate benefits both from foresight of the predictive and the accuracy of the local estimate to meet conditions (a) and (b).

%Because $f_{\text{post}}$ is a distribution over functions, it does not only give us the maximum likelihood predictive friction (the mean), but also a measure of the associated uncertainty in the estimate, at each position $s$. 

%This richer representation allows the motion planner to make more informed decisions about allowed tire forces in critical situations. 

% a word on tuning parameters?
% *length scale
% s locall threshold

% result
\section{Results and Discussion}
\label{sec:results}

% intro
In this section, we evaluate the impact of friction estimation performance in the context of traction adaptive motion planning. In a simulated environment, we emulate (i.e., replicate the output of) state of the art predictive and local friction estimation algorithms in terms of their availability, foresight and accuracy. We then compare local only, predictive only and fused configurations to ground truth, in terms of the resulting motion behavior.

%under four friction estimation configurations, including the proposed approach of fusing heterogeneous friction estimates through GP regression.

\subsection{Delimitations and Experimental Setup}
\label{sec:results:setup}

% Motivate emulation of friction estimation algos
For the purposes of the evaluation we emulate rather than implement state of the art friction estimation algorithms. This choice is motivated by practical as well as methodological reasons. 
The focus of this paper is to evaluate various strategies for handling uncertain and not always available friction estimates, in the context of traction adaptive motion planning. 
%This paper concerns motion planning in critical situations, and in that context, how best to utilize uncertain and not always available friction estimates. 
Given this purpose, implementing our own and/or a selection of state-of-the-art methods for friction estimation would a) be a substantial time investment and b) introduce an additional risk of biasing the resulting motion behavior through stochastic variations or inperfections in the implementation. Instead, we have opted for parameterizing the performance of state of the art estimation algorithms, and emulating them in a simulation environment. This way we can draw conclusions based on the fundamental properties of the methods. %, and design fusion methods that achieve the desired behavior.  
% mention simulator
Simulations are performed using an extended version of the Gazebo based FSSIM simulator \cite{kabzan2020amz}, that have been modified and tuned to represent the REVERE lab's Volvo FH16 test vehicle, which was used in the experimental evaluation of \cite{svensson2020traction}.

Based on the literature summarized in Section \ref{sec:intro}, we select parameter specifications to represent current state-of-the-art performance in local and predictive road friction estimation in Table \ref{tab:mu_est_methods_perfparams}.
% explain vars in table
Variables $e_p$ and $e_l$ denote estimation errors for the predictive and local methods respectively. The parameter $\lambda_t \in [0,1]$ denotes the momentary utilization of available tire force. 
\begin{table}[h] 
    \centering
    \begin{tabular}{c|c|c}
         & \bfseries Accuracy & \bfseries Availability\\
        \hline
        % values from "Quantification of Excitation Required for Accurate Friction Estimation"
        Local (L) & $\max\{|e_l|\} = 0.025$, \cite{prokevs2016quantification} & \begin{tabular}{@{}c@{}} $\lambda_t > 0.5$, \cite{prokevs2016quantification}  \\ 
        %(convg. time 0.06s \cite{prokevs2016quantification}) 
        \end{tabular} \\
        Predictive (P) & \begin{tabular}{@{}c@{}} $\max\{|e_p|\} \approx 0.1 ~\text{to}~ 0.3$ \\ classes: dry, wet, snow/ice \cite{sohini2018predfrictionest}  \end{tabular}   & \begin{tabular}{@{}c@{}} $s \in \{0,ds,\dots, s_f\}$, \\ $s_f = 50$m \end{tabular} \\ % (3 classes:dry asphalt, wet asphalt, snow/ice)
        \hline
    \end{tabular}
    \vspace{0.2cm}
    \caption{Selected performance parameters for state of the art friction estimation methods}
    \label{tab:mu_est_methods_perfparams}
\end{table}

\subsection{Evaluated Configurations}
We compare four different friction estimation configurations: Ground truth (GT), Local Only (L), Predictive only (P), and Fused (F). Here follows a description of each configuration.

% GT 
The first configuration, GT, is a non-realistic benchmark, representing the optimal behavior on the current road surface. The traction adaptive motion planner/controller is provided the true friction values from the track ahead of the vehicle, i.e.,
\begin{equation*}
    \hat{\mu}(s) = \mu_{\text{gt}}(s) \quad \text{for all} \quad s \in \{0,ds,\dots, s_f\}.
\end{equation*}
% L
In the local only configuration, L, we emulate a local friction estimate being propagated over the whole prediction horizon, i.e., $\hat{\mu}(s) = \hat{\mu}_l$ for all $s \in \{0,ds,\dots, s_f\}$, with
\begin{equation*}
    \hat{\mu}_l = -\max\{|e_l|\} + 
    \begin{cases}
        \mu_{\text{gt}}(0) + e_l  ~~  \quad \text{if local available} \\
        \hat{\mu}_0 \quad \quad \quad \quad \quad  \text{otherwise}
    \end{cases}       
\end{equation*}
The estimate is conservative in the sense that we subtract the worst case error before passing the estimate to the planner, so as to prevent over-estimation. If the local estimate is unavailable, the latest available estimate $\hat{\mu}_0$ is used. 

% P
In the predictive only configuration, P, the planner/controller is provided a conservative predictive friction estimate. The estimate is conservative in the sense that it takes the minimum value of the surface class associated with the point $s$ 
\begin{equation*}
  \hat{\mu}(s) =
    \begin{cases}
      0.6 & \text{if} \quad 0.6 < \mu_{\text{gt}}(s) \quad \quad \quad ~~\text{class: dry} \\
      0.4 & \text{if} \quad 0.4 \leq \mu_{\text{gt}}(s) < 0.6    \quad \text{class: wet} \\
      0.1 & \text{if} \quad 0.1 \leq \mu_{\text{gt}}(s) < 0.4    \quad \text{class: snow/ice.}
    \end{cases}       
\end{equation*}

% measurements in F (todo move to method ?)
In the fused configuration F, we emulate both local and predictive friction estimation such that the combined information from configurations 2 and 3 is made available. 
\begin{align*}
  &\hat{\mu}'(s), m'(s) = \\
    &~~~\begin{cases}
      0.8, 0.2   & \text{if} \quad 0.6 < \mu_{\text{gt}}(s) \quad \quad \quad ~~\text{class: dry} \\
      0.5, 0.1   & \text{if} \quad 0.4 \leq \mu_{\text{gt}}(s) < 0.6    \quad \text{class: wet} \\
      0.25, 0.15 & \text{if} \quad 0.1 \leq \mu_{\text{gt}}(s) < 0.4    \quad \text{class: snow/ice} \\
      \mu_{\text{gt}}(0) + e_l,& \max\{|e_l|\} \quad \text{if local available \& } s<s_l 
    \end{cases}       
\end{align*}
For the predictive estimates, instead of directly providing the lowest value in each class, we extract the mean values $\hat{\mu}'(s)$ along with the margins of error $m'(s)$. With this input data, we obtain a fused estimate $\hat{\mu}(s)$ using the method described in Section \ref{sec:method}.

% Motvation of the cases: 
Notice that configurations L, P and F, all set conservative estimates with respect to the available information. As mentioned in Section \ref{sec:method}, this is a generally applied principle in using online friction estimation for vehicle control. 

% describe scenarios + WC
We evaluate the four configurations in two simulated critical situations; Scenario 1, Turn at low $\mu$, and Scenario 2, Collision avoidance at high $\mu$. For both scenarios, we set adversarial worst case friction estimation errors, i.e., worst case over-estimation in the low $\mu$ turn scenario (where inability to realize large planned tire forces will lead to accident) and worst case under-estimation in the collision avoidance scenario (where overly restrictive tire force constraints will lead to accident). 

% bridge to exp1

%  and third, a challenging gauntlet of multiple collision avoidance maneuvers on variable $\mu$ conditions. 
% mention that scenarios are worst case? (different for turn and ca point to tamp paper)

% \begin{figure}[t]
%     \centering
%     \includegraphics[width=0.30\textwidth]{figs/cam_fric_mapping.pdf}\\
%     \caption{Example mapping of classes in camera based road surface classification to friction coefficient}
%     \label{fig:cam_fric_mapping}
% \end{figure}

\subsection{Scenario 1 - Turn at low $\mu$}
\label{sec:results:scenario_1}

\begin{figure*}[!]
    \captionsetup[subfigure]{}
    \centering
    \subfloat[Local only, $s=0$m]{%
        \includegraphics[width=0.24\textwidth]{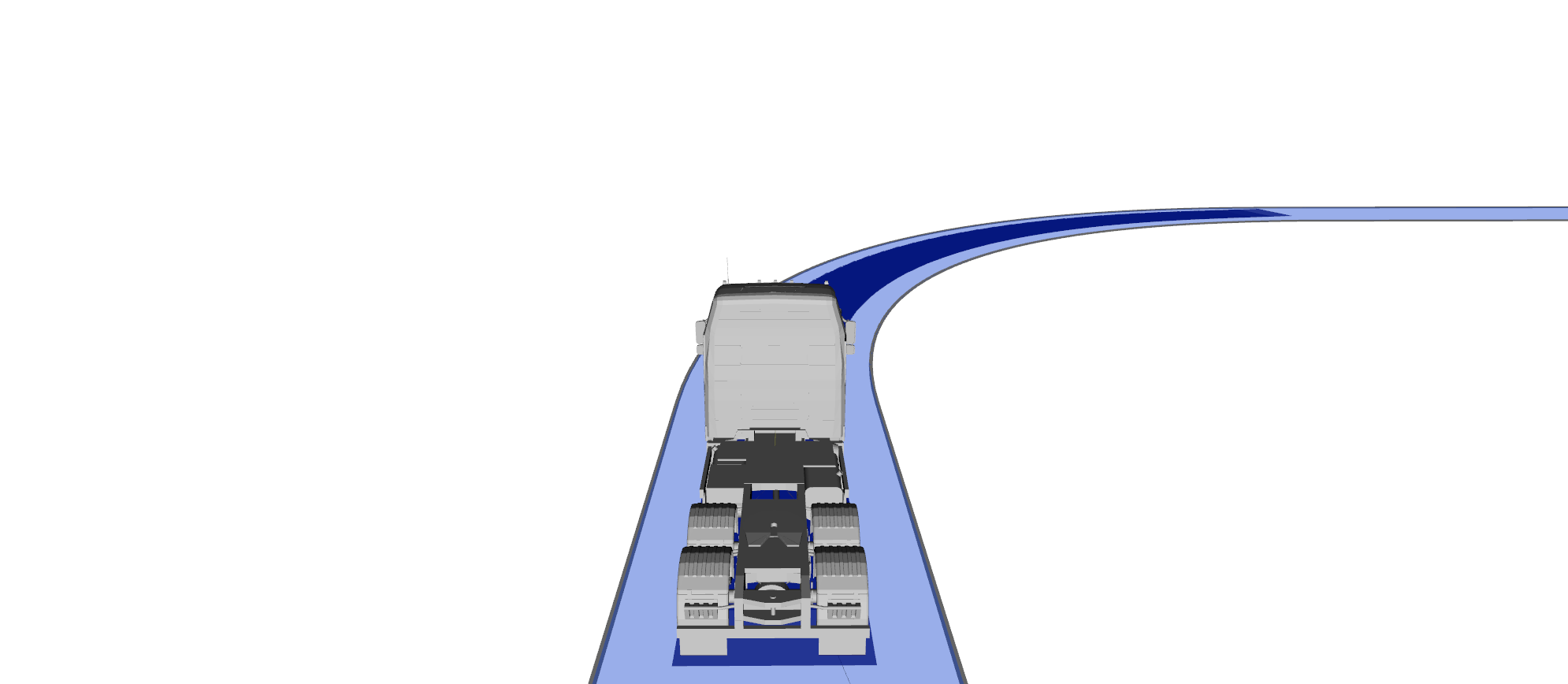}
        \label{fig:lowmuturn:local_wce_0}
    }
    \subfloat[Local only, $s=15$m]{%
        \includegraphics[width=0.24\textwidth]{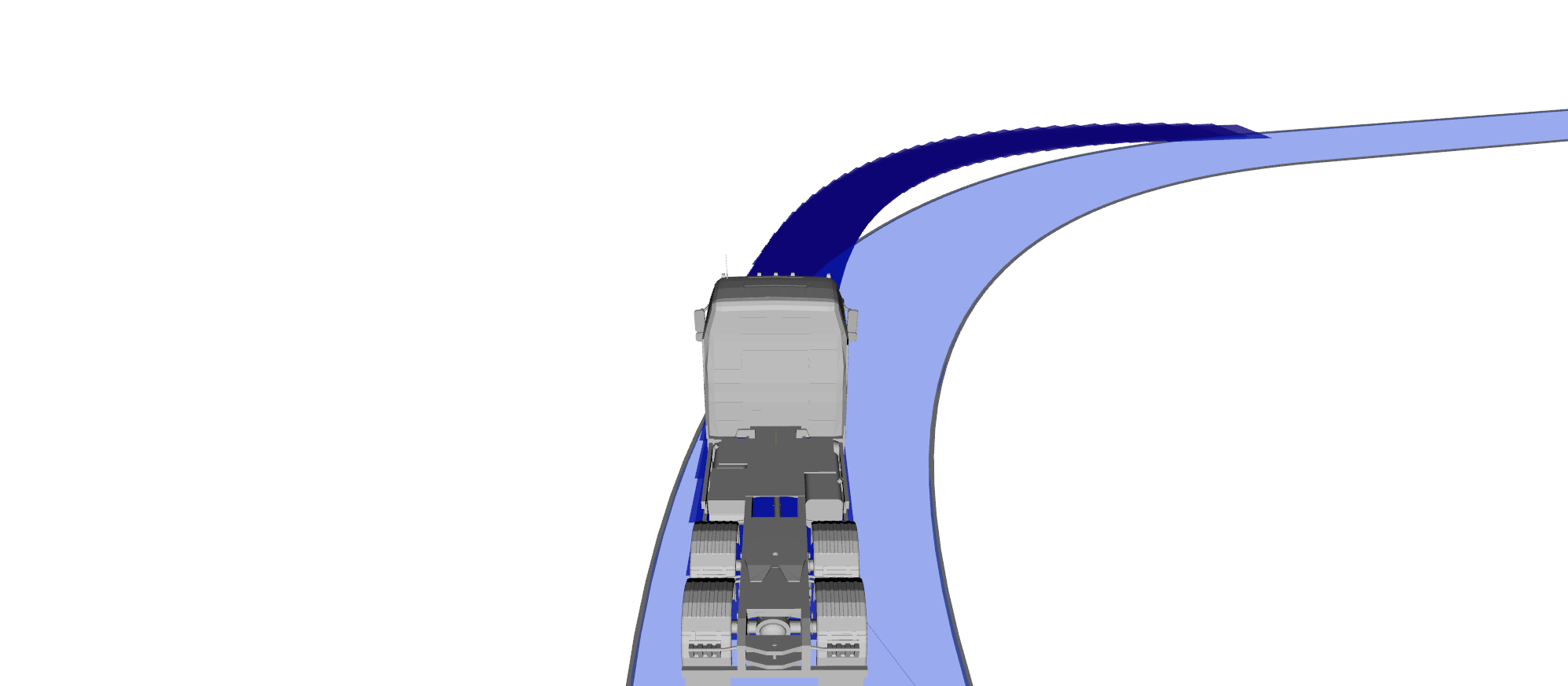}
        \label{fig:lowmuturn:local_wce_1}
    }
    \subfloat[Local only, $s=30$m]{%
        \includegraphics[width=0.24\textwidth]{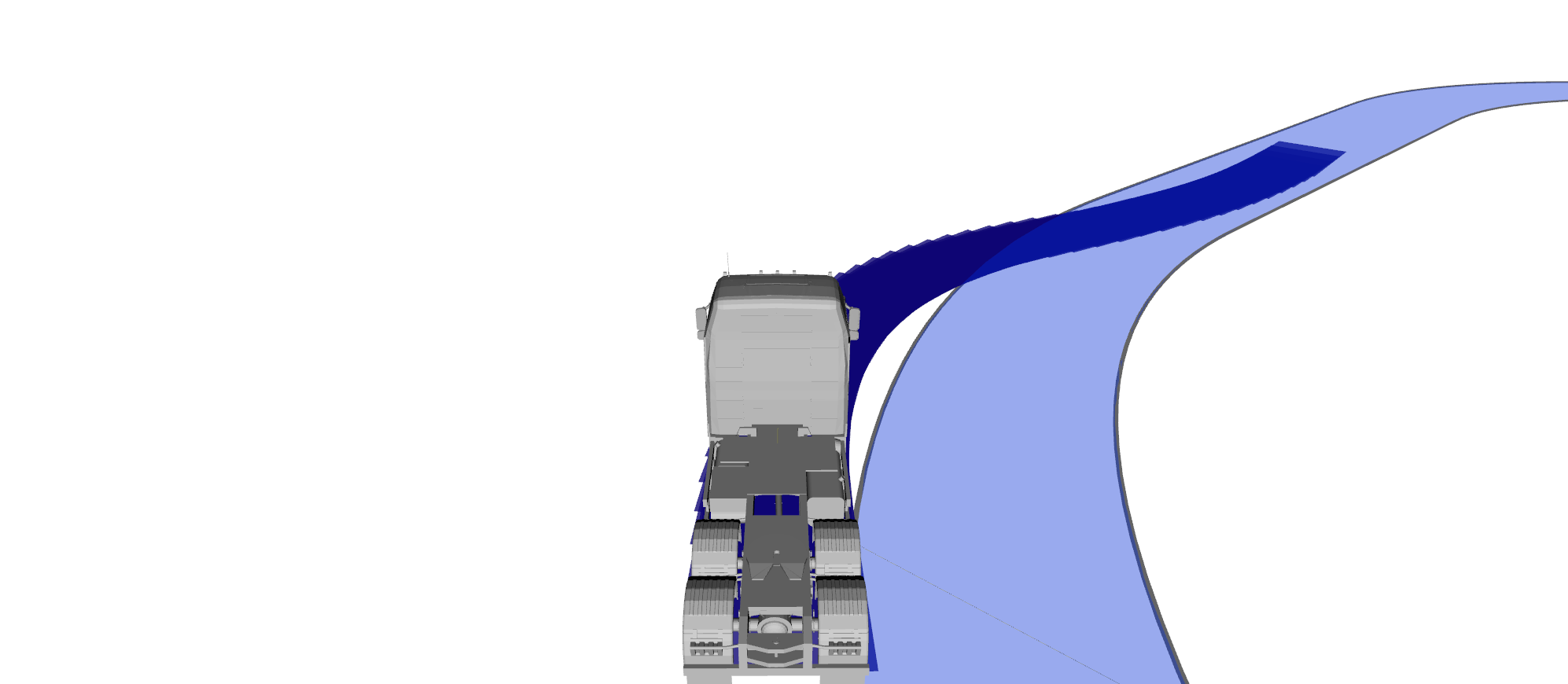}
        \label{fig:lowmuturn:local_wce_2}
    }
    \subfloat[Local only, $s=55$m]{%
        \includegraphics[width=0.24\textwidth]{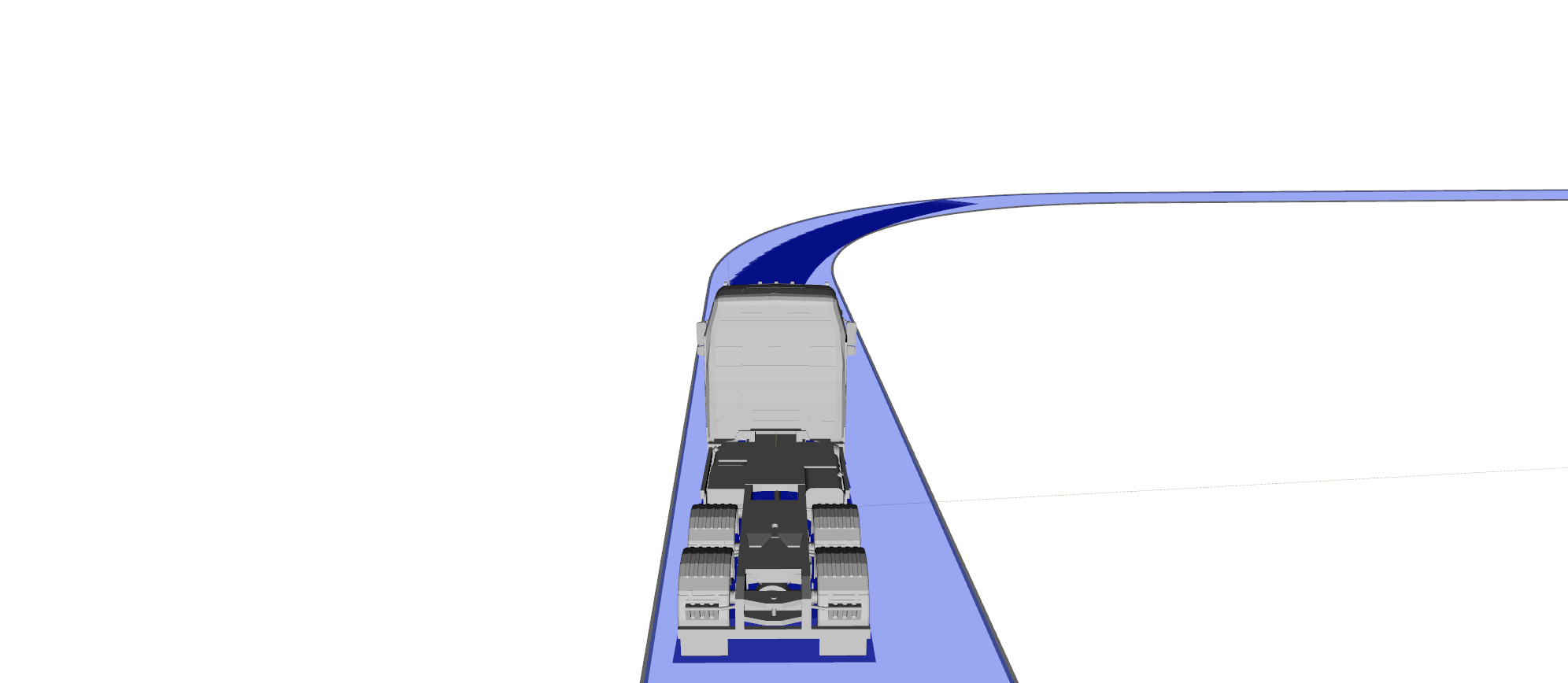}
        \label{fig:lowmuturn:local_wce_3}
    }\\
    \subfloat[Fused, $s=0$m]{%
        \includegraphics[width=0.24\textwidth]{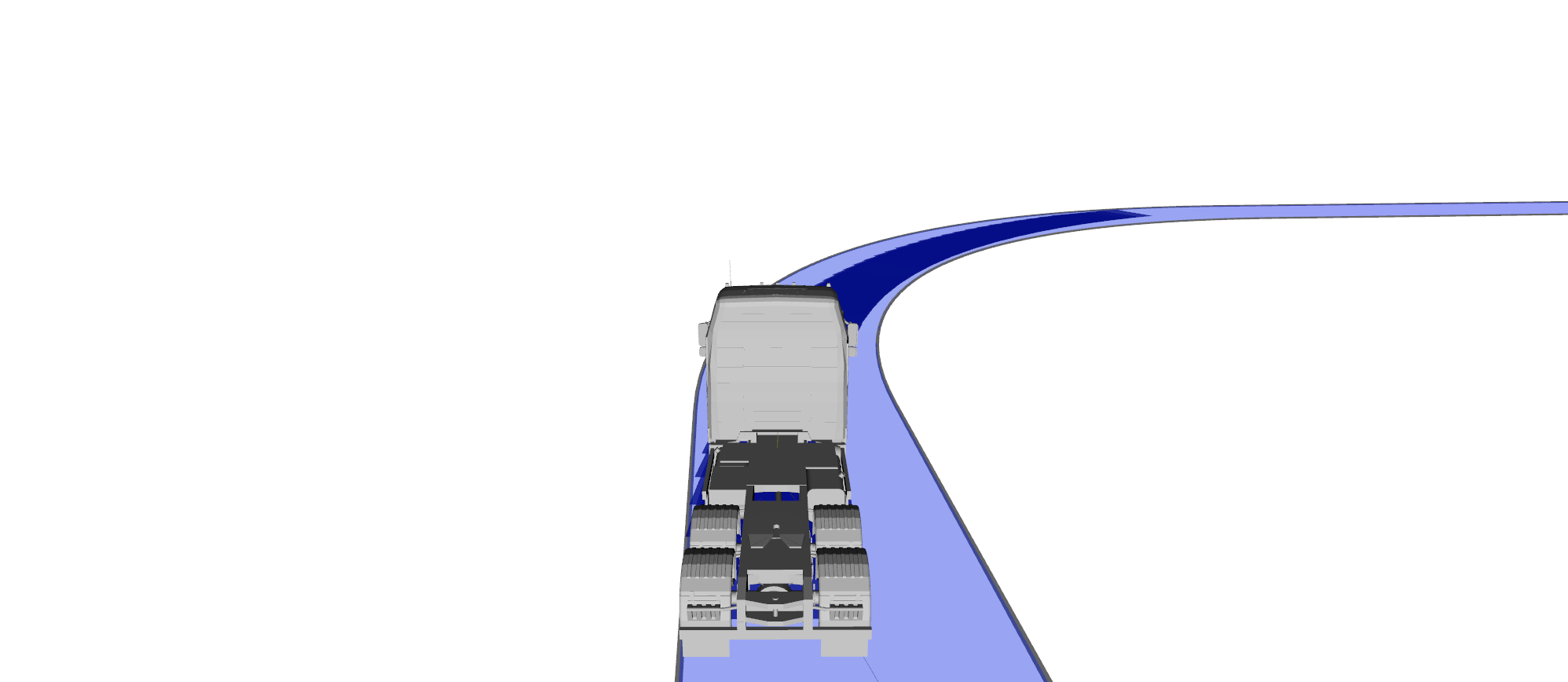}
        \label{fig:lowmuturn:gp_wce_0}
    }
    \subfloat[Fused, $s=15$m]{%
        \includegraphics[width=0.24\textwidth]{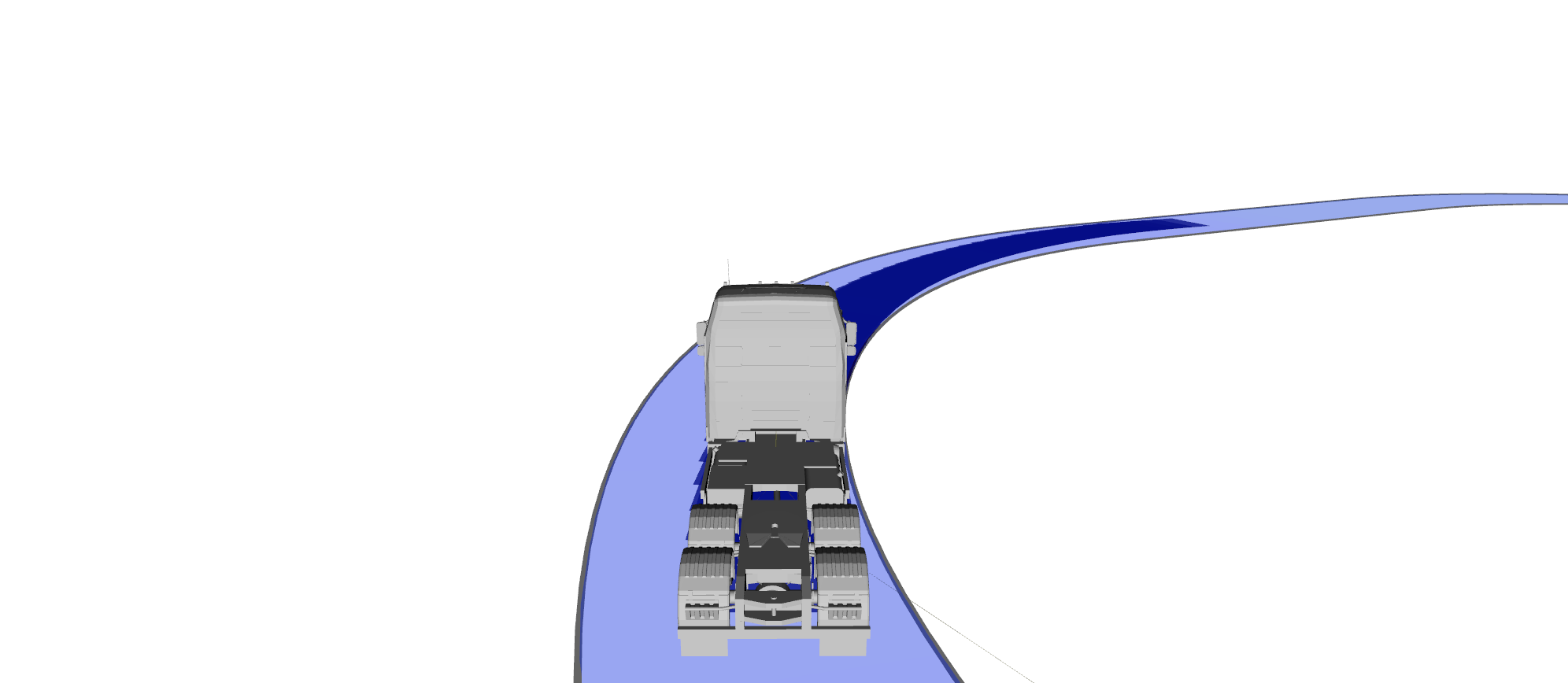}
        \label{fig:lowmuturn:gp_wce_1}
    }
    \subfloat[Fused, $s=30$m]{%
        \includegraphics[width=0.24\textwidth]{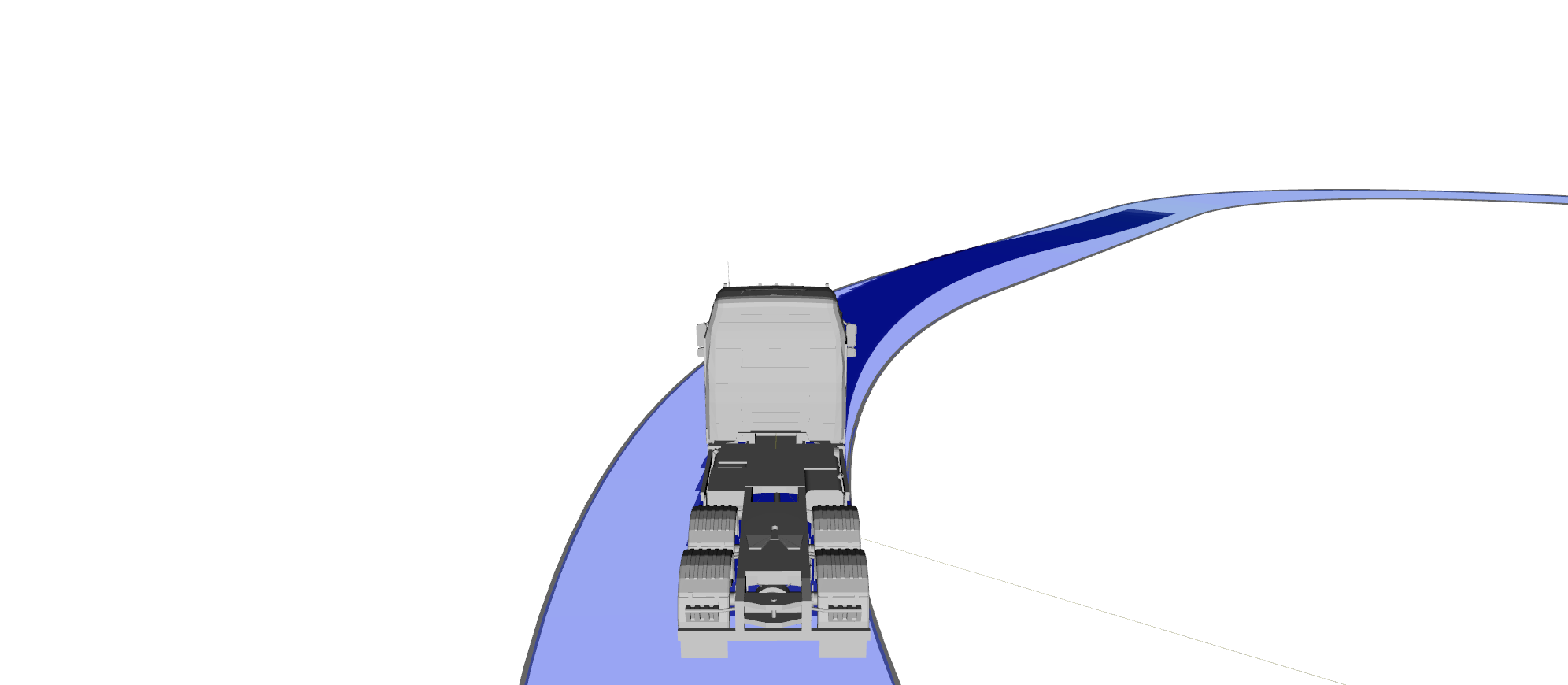}
        \label{fig:lowmuturn:gp_wce_2}
    }
    \subfloat[Fused, $s=55$m]{%
        \includegraphics[width=0.24\textwidth]{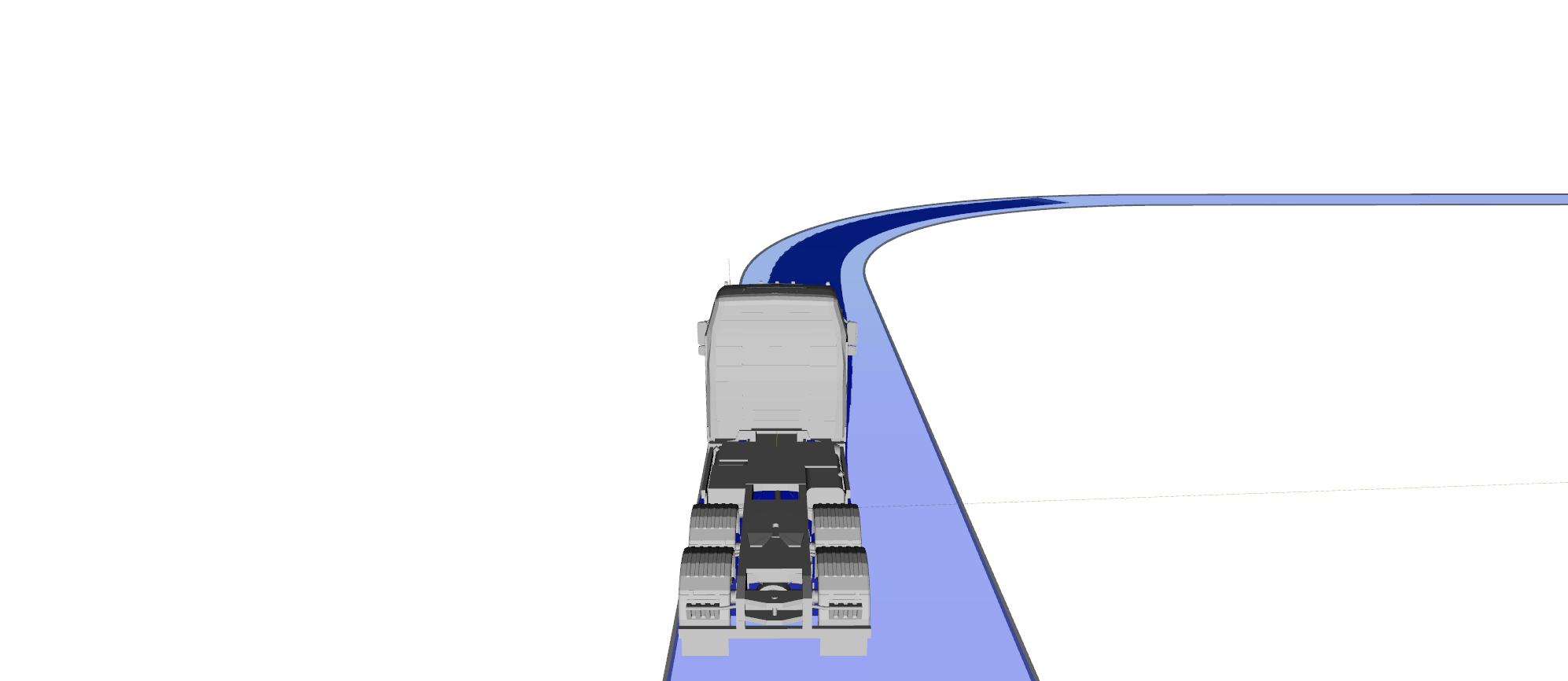}
        \label{fig:lowmuturn:gp_wce_3}
    }\\
    \subfloat[Distance to lane center]{%
        \includegraphics[width=0.45\textwidth]{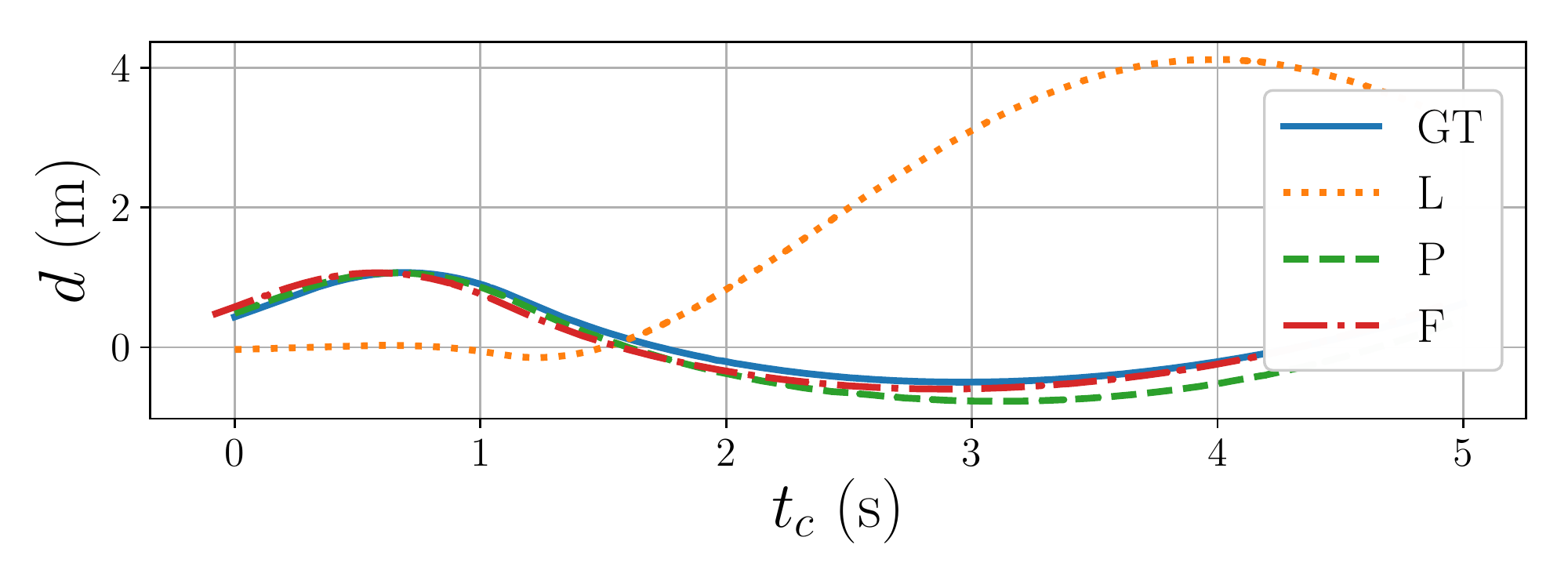}
        \label{fig:lowmuturn:d}
    }
    \subfloat[Velocity]{%
        \includegraphics[width=0.45\textwidth]{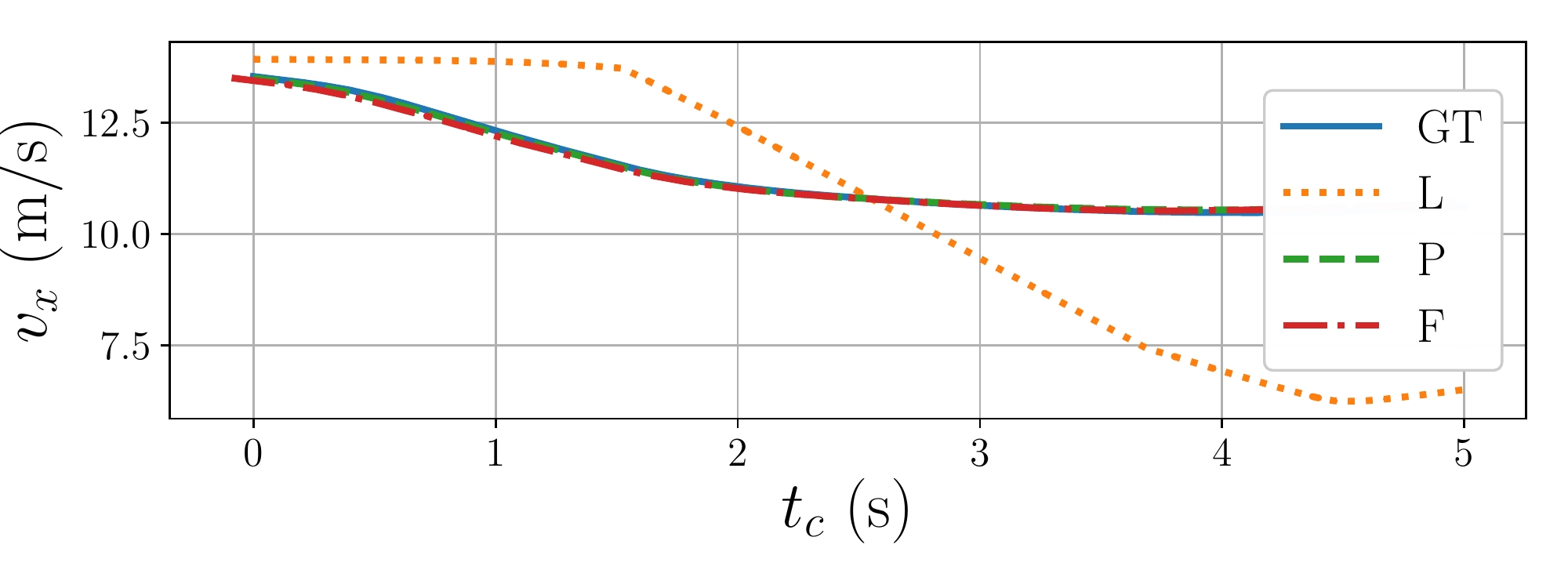}
        \label{fig:lowmuturn:vx}
    } \\
    \subfloat[Planned front tire forces at $t_c=0$]{%
        \includegraphics[width=0.45\textwidth]{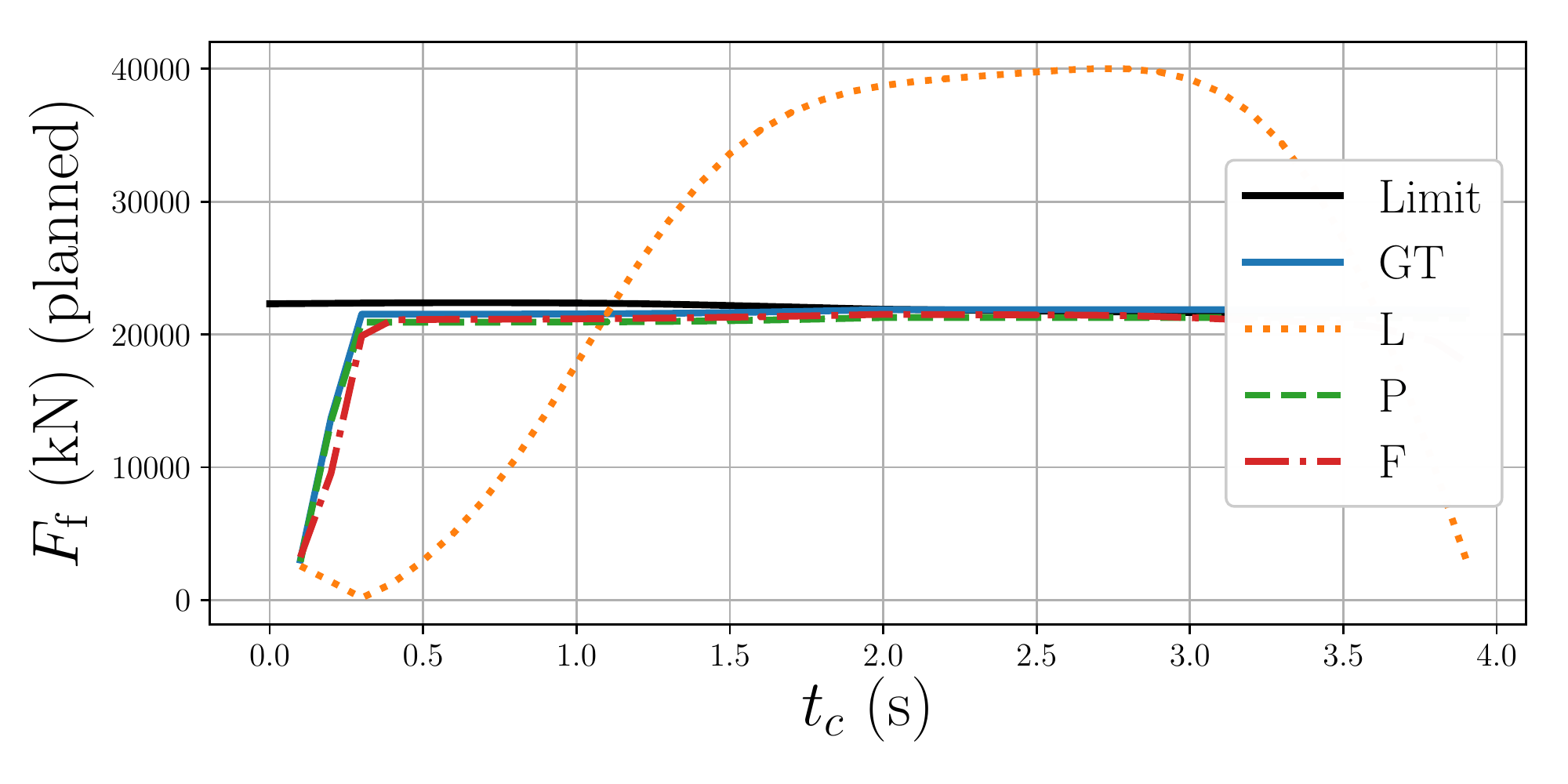}
        \label{fig:lowmuturn:Ff}
    }
    \subfloat[Merged predictive estimate with uncertainty at $t_c = 0.5$]{%
        \includegraphics[width=0.45\textwidth]{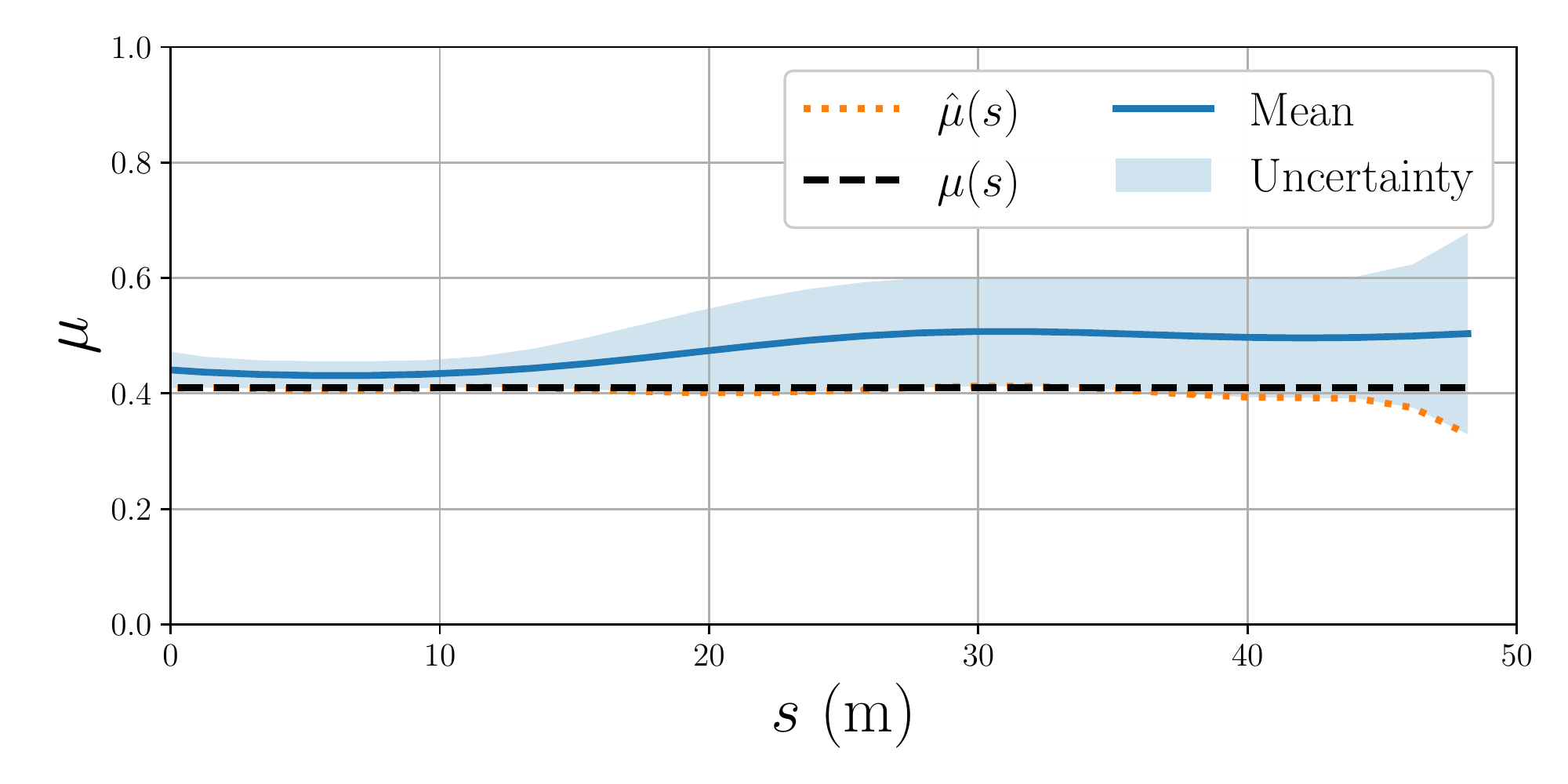}
        \label{fig:lowmuturn:gp}
    }
    \centering
    \caption{Simulation results for the first critical scenario - a 90 degree turn at low $\mu$ with worst case over-estimation of friction. Subfigures \ref{fig:lowmuturn:local_wce_0} through \ref{fig:lowmuturn:gp_wce_3} show snapshots from the online visualization of the planned motion. The first row (\ref{fig:lowmuturn:local_wce_0} - \ref{fig:lowmuturn:local_wce_3}) correspond to the local only configuration and the second row (\ref{fig:lowmuturn:gp_wce_0} - \ref{fig:lowmuturn:gp_wce_3}) correspond to the proposed fused configuration. 
    Subfigures \ref{fig:lowmuturn:d} and \ref{fig:lowmuturn:vx} show the vehicle's distance to the lane center and velocity for the four configurations. 
    Subfigure \ref{fig:lowmuturn:Ff} show planned front tire forces at $t_c=0$s, $s=0$m and the prevailing force limits.
    Subfigure \ref{fig:lowmuturn:gp} shows the corresponding friction estimate for the fused configuration, including uncertainty.}
    \label{fig:lowmuturn:full}
\end{figure*}

At the start of the first scenario, $t_c=0$, $s=0$, the vehicle is approaching a 90 degree turn at a velocity of $12$m/s. Its objective is to maintain velocity and stay at the lane center. The traction in the corner is locally reduced, $\mu(s) = 0.4$ for $s \geq 0$m, from previously good conditions $\mu(s) = 0.8$ for $s < 0$m.
Fig.~\ref{fig:lowmuturn:full}, shows an analysis of the generated motion behavior for the four friction estimation configurations.

% point out that L is different, others are same
At first glance at Figures \ref{fig:lowmuturn:d} and \ref{fig:lowmuturn:vx}, we see that configurations GT, P and F generate similar behavior while L shows distinctly different behavior. 
% describe L
Under the L configuration, the vehicle enters the turn at the center of the lane, and plans to stay at the center throughout the turn, Fig.~\ref{fig:lowmuturn:local_wce_0}. Upon entering the corner, Fig.~\ref{fig:lowmuturn:local_wce_1}, the vehicle starts slipping toward the outside of the lane, deviating from the planned motion. At the middle of the turn, Fig.~\ref{fig:lowmuturn:local_wce_2}, the vehicle has slipped well into the opposing lane, at risk of colliding with oncoming traffic. By now however, the vehicle has managed to reduce its speed such that it is able to recover to the lane, Fig.~\ref{fig:lowmuturn:local_wce_3}. 

% explain why
The reason behind this undesired motion behavior is the that the L configuration relies entirely on a local friction estimation, which is only available when tire force utilization $\lambda > 0.5$. While travelling on the straight road before the turn at constant velocity, $\lambda << 0.5$. Thus tire force constraints are set based on the latest available friction estimate, here $\hat{\mu}_0 = 0.8$. As is shown in Fig.~\ref{fig:lowmuturn:Ff}, the planned tire forces at $t_c=0$ for configuration L exceeds the prevailing tire force limit by a wide margin.  
The moment the vehicle starts turning, Fig.~\ref{fig:lowmuturn:local_wce_1}, the planner reduces the tire force constraints and adjusts the planned motion, but at this time, it is physically impossible to avoid veering out into the opposing lane. 

% describe other three
In contrast, the other three configurations GT, P and F  exhibit the behavior represented by F in Subfigures \ref{fig:lowmuturn:gp_wce_0} - \ref{fig:lowmuturn:gp_wce_3}. Before entering the turn, Fig.~\ref{fig:lowmuturn:gp_wce_0}, the vehicle has positioned itself on the outside of the lane, and plans to cut across to the inside of the corner by the middle of the turn and then back to the outside of the lane by the end of the turn, while slightly reducing speed. As is shown in Figures \ref{fig:lowmuturn:gp_wce_1}, \ref{fig:lowmuturn:gp_wce_2} and \ref{fig:lowmuturn:gp_wce_3}, the vehicle does not deviate from the planned motions, and the updated plans are consistent with the previous ones.   

% explain how predictive conservative fixes the problem
The reason behind the difference in behavior is the foresight provided by the predictive friction estimate. This in combination with a conservative selection of tire force limits ensures that the planned tire forces do not exceed the physical limits in either of the three configurations \ref{fig:lowmuturn:Ff}. The traction adaptive planner/controller realizes the reduced tire forces by a coordinated speed reduction and lateral motion in the lane as shown in Figures \ref{fig:lowmuturn:d} and \ref{fig:lowmuturn:vx}. 

% comment on GP merge
For the F configuration, because the maneuver is initiated earlier, the local friction estimate becomes available earlier compared to the L configuration. Fig.~\ref{fig:lowmuturn:gp} shows the fused friction estimate at $t_c=0.5$. 
%The conservative estimate is selected as the lower bound of the uncertainty region of the posterior distribution. 
% comment on the fact that GT, P and F are pretty much the same
In this scenario, where the emulated friction estimation error equals the worst case over-estimation, the fused predictive estimate more or less coincides with the ground truth friction value. The same is the case for P. For this reason, the three configurations show very similar planned tire forces in Fig.~\ref{fig:lowmuturn:Ff}, and consequently similar motion behavior. In cases where the estimation errors are below the worst case over-estimation, configurations P and F will generate more conservative planned forces, through keeping the same lateral motion pattern, but reducing velocity even more before the turn. Between the two, configuration F will be less conservative due to the higher accuracy of the local estimate. Next, we evaluate the effect of this phenomenon in collision avoidance.

\subsection{Scenario 2 - CA at high $\mu$}
\label{sec:results:scenario_2}

\begin{figure*}[!]
    \captionsetup[subfigure]{}
    \centering
    \subfloat[Predictive only, $s=0$m]{%
        \includegraphics[width=0.24\textwidth]{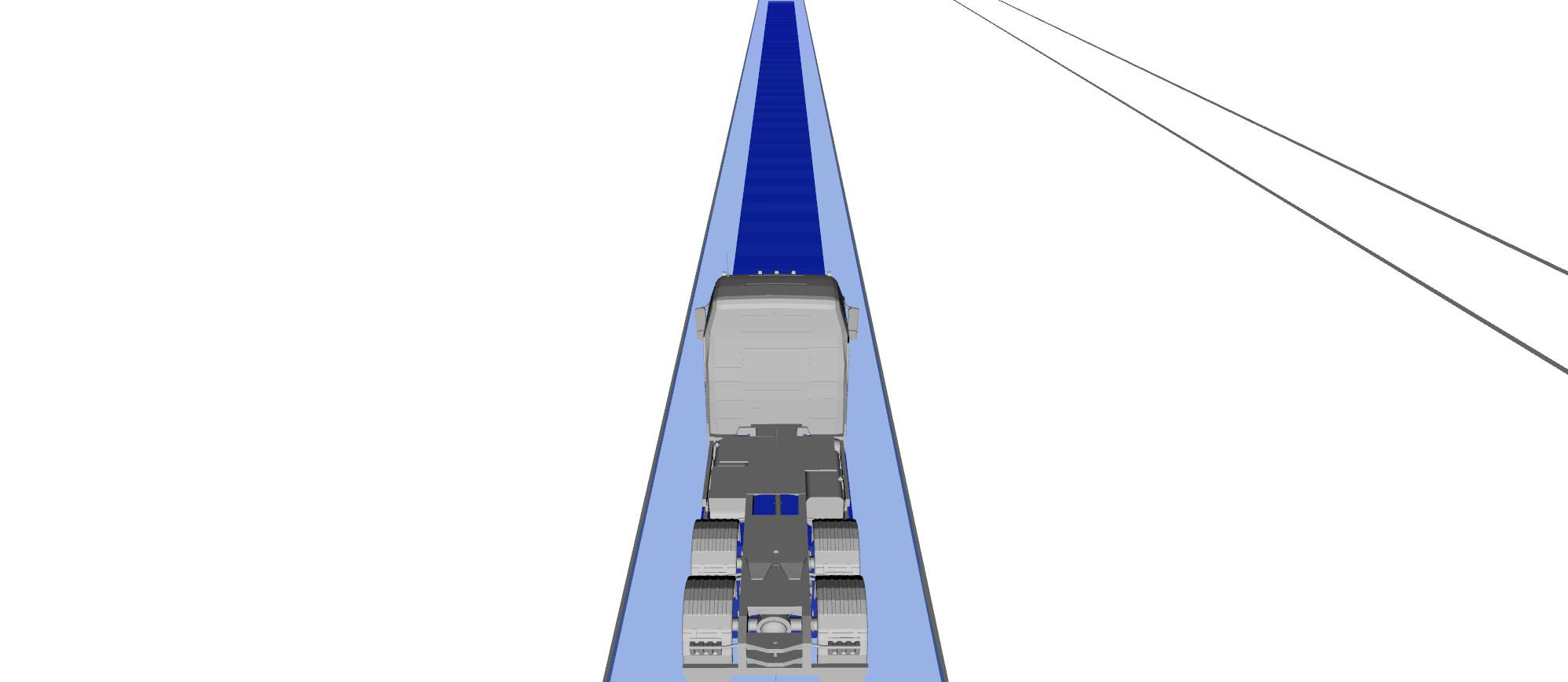}
        \label{fig:collavoid:pred_wce_0}
    }
    \subfloat[Predictive only, $s=2$m]{%
        \includegraphics[width=0.24\textwidth]{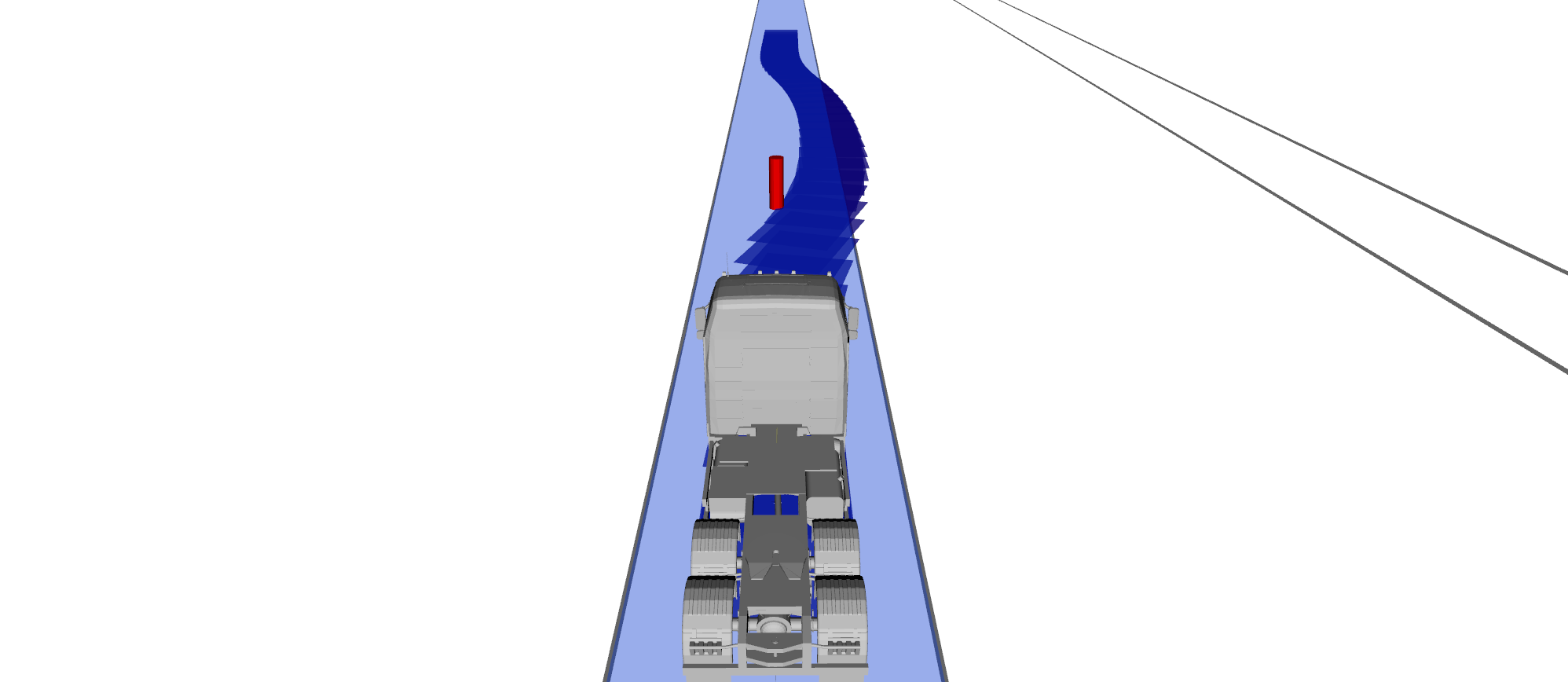}
        \label{fig:collavoid:pred_wce_1}
    }
    \subfloat[Predictive only, $s=18$m, \\  \textbf{(collision)}]{%
        \includegraphics[width=0.24\textwidth]{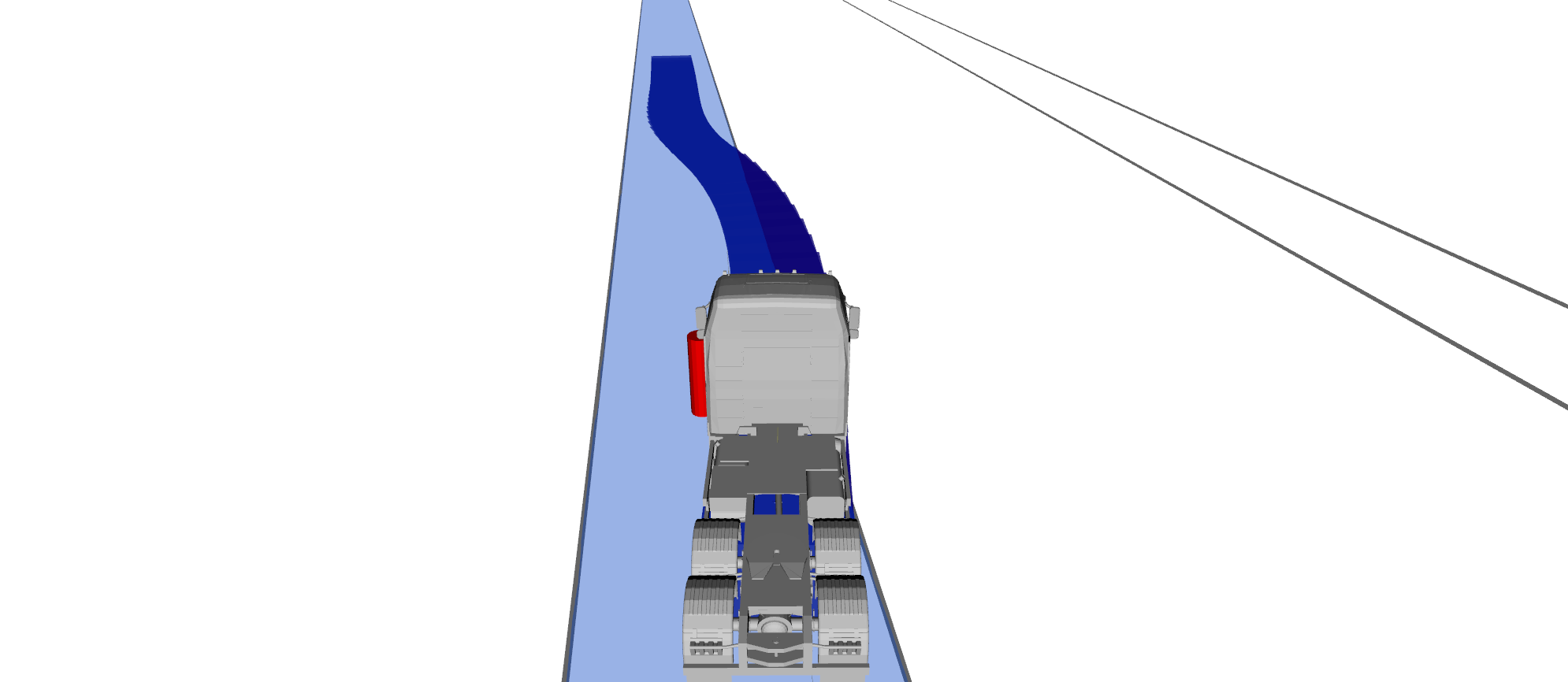}
        \label{fig:collavoid:pred_wce_2}
    }
    \subfloat{%
        \includegraphics[width=0.24\textwidth]{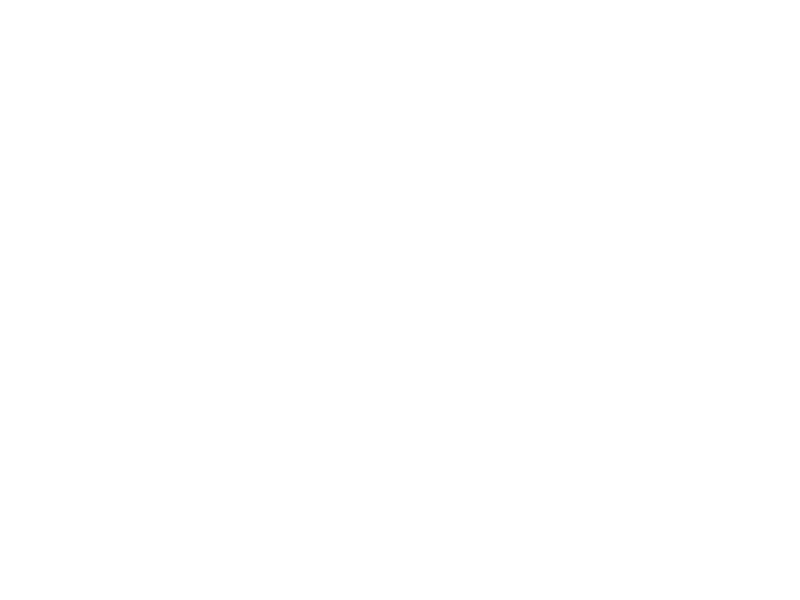}
        \label{fig:collavoid:pred_wce_3}
    }\\
    \addtocounter{subfigure}{-1} % fixes index gap
    \subfloat[Fused, $s=0$m]{%
        \includegraphics[width=0.24\textwidth]{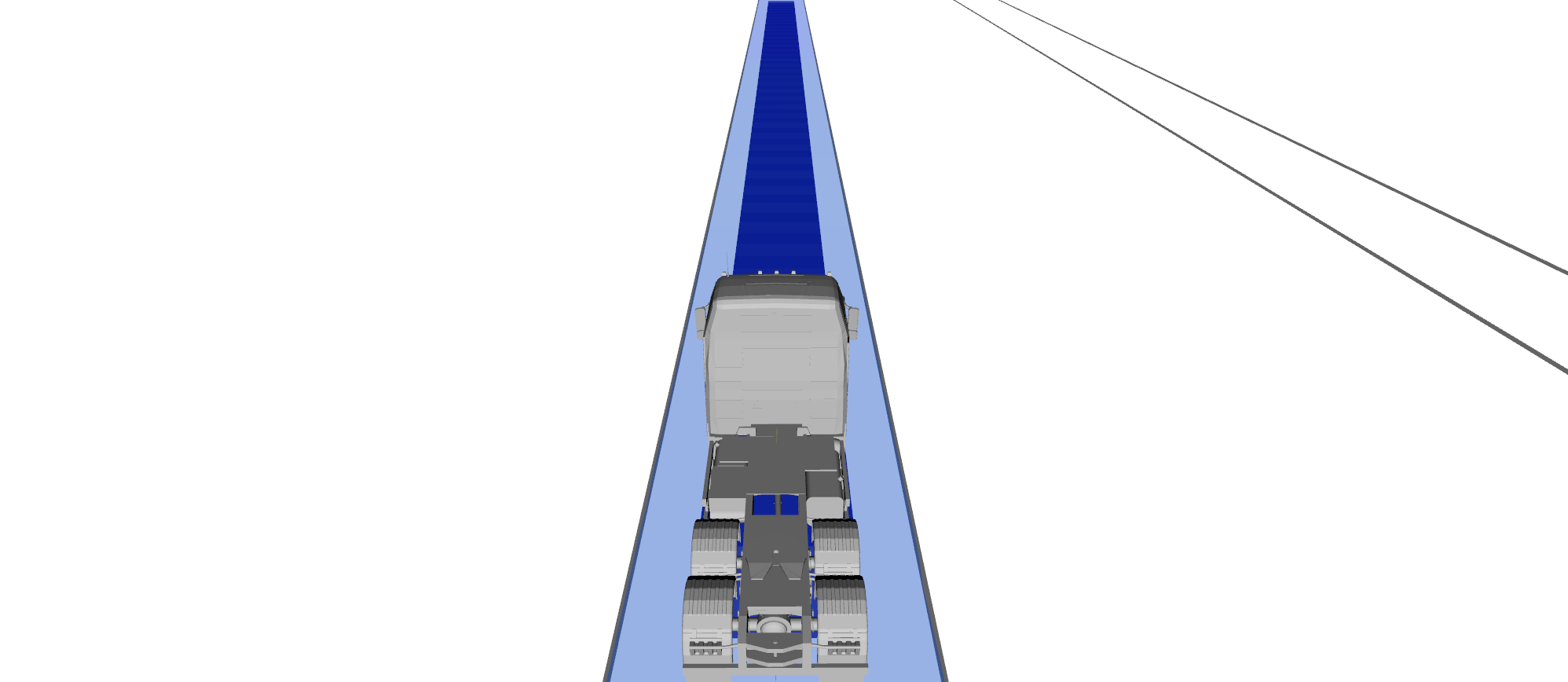}
        \label{fig:collavoid:gp_wce_0}
    }
    \subfloat[Fused, $s=2$m]{%
        \includegraphics[width=0.24\textwidth]{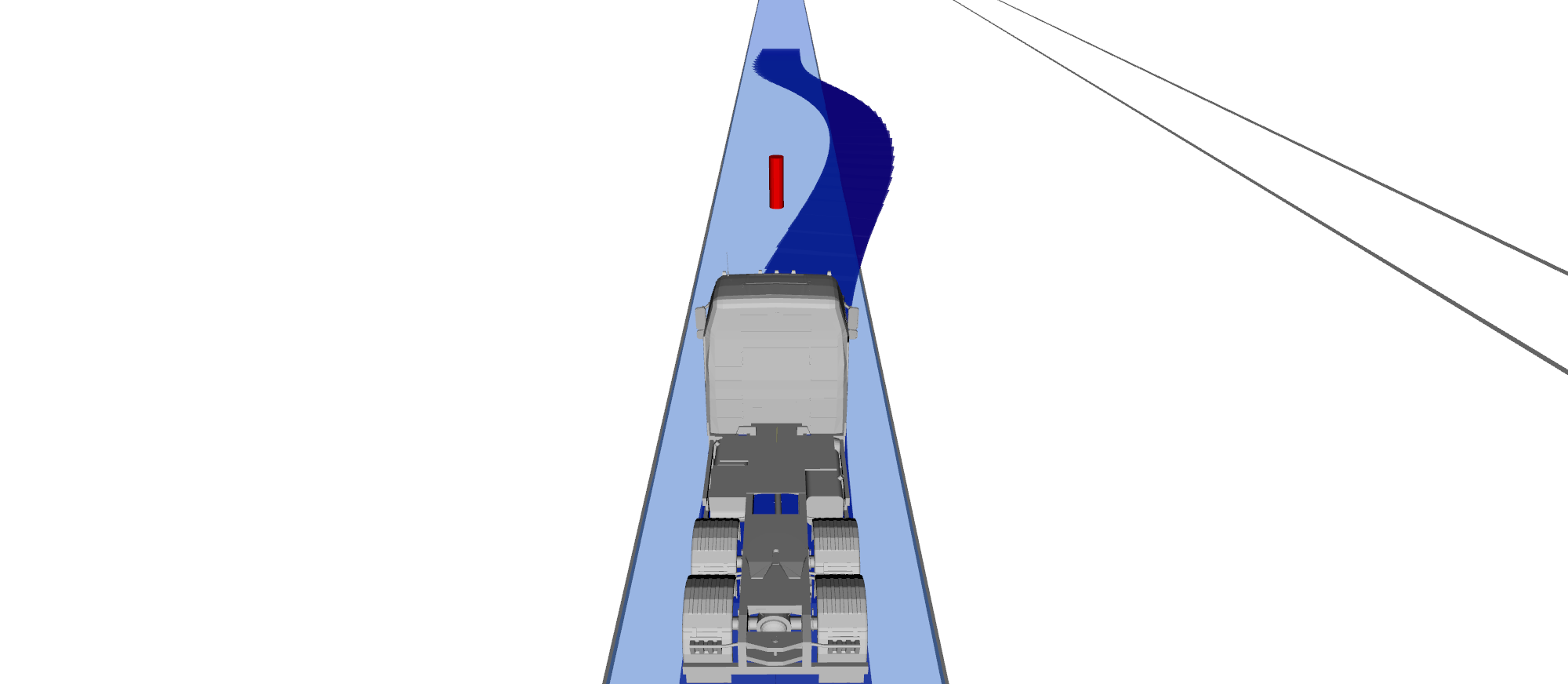}
        \label{fig:collavoid:gp_wce_1}
    }
    \subfloat[Fused, $s=18$m]{%
        \includegraphics[width=0.24\textwidth]{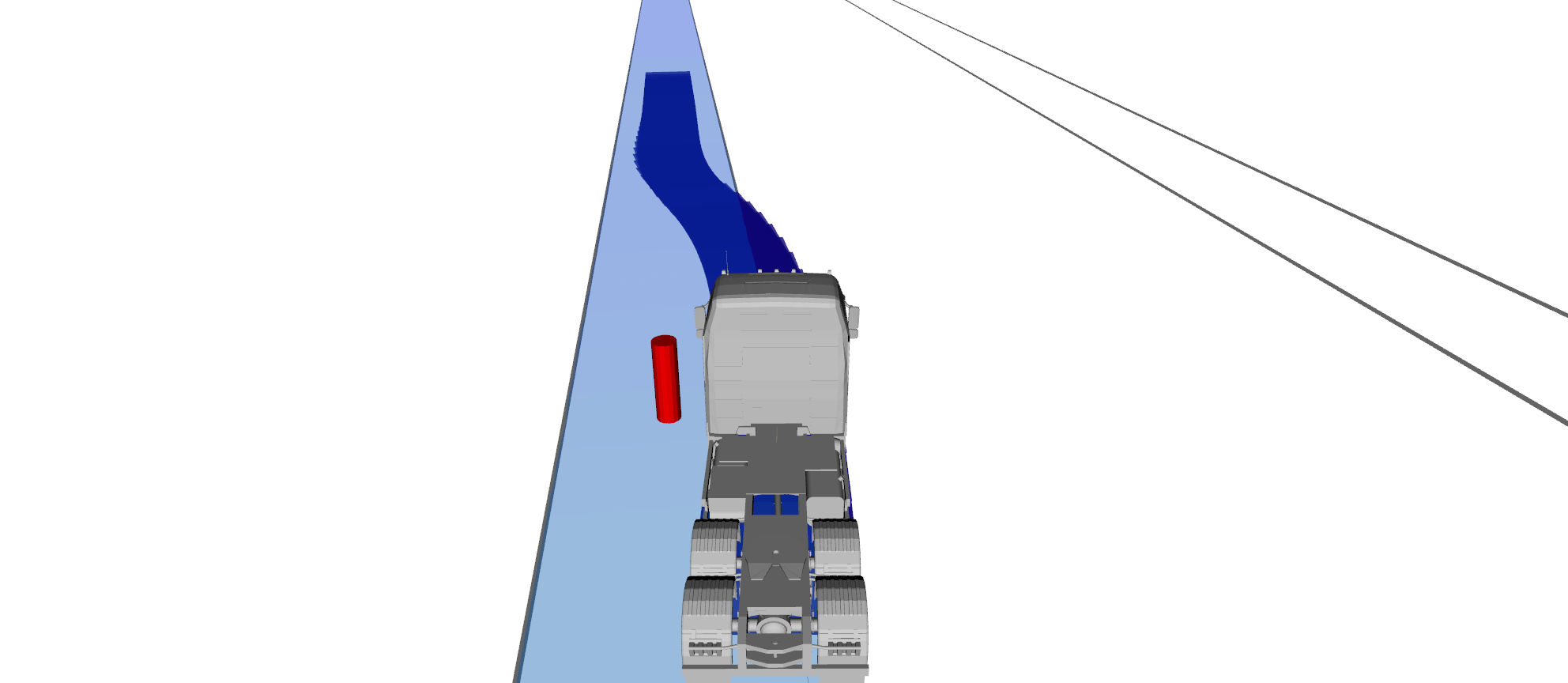}
        \label{fig:collavoid:gp_wce_2}
    }
    \subfloat[Fused, $s=40$m]{%
        \includegraphics[width=0.24\textwidth]{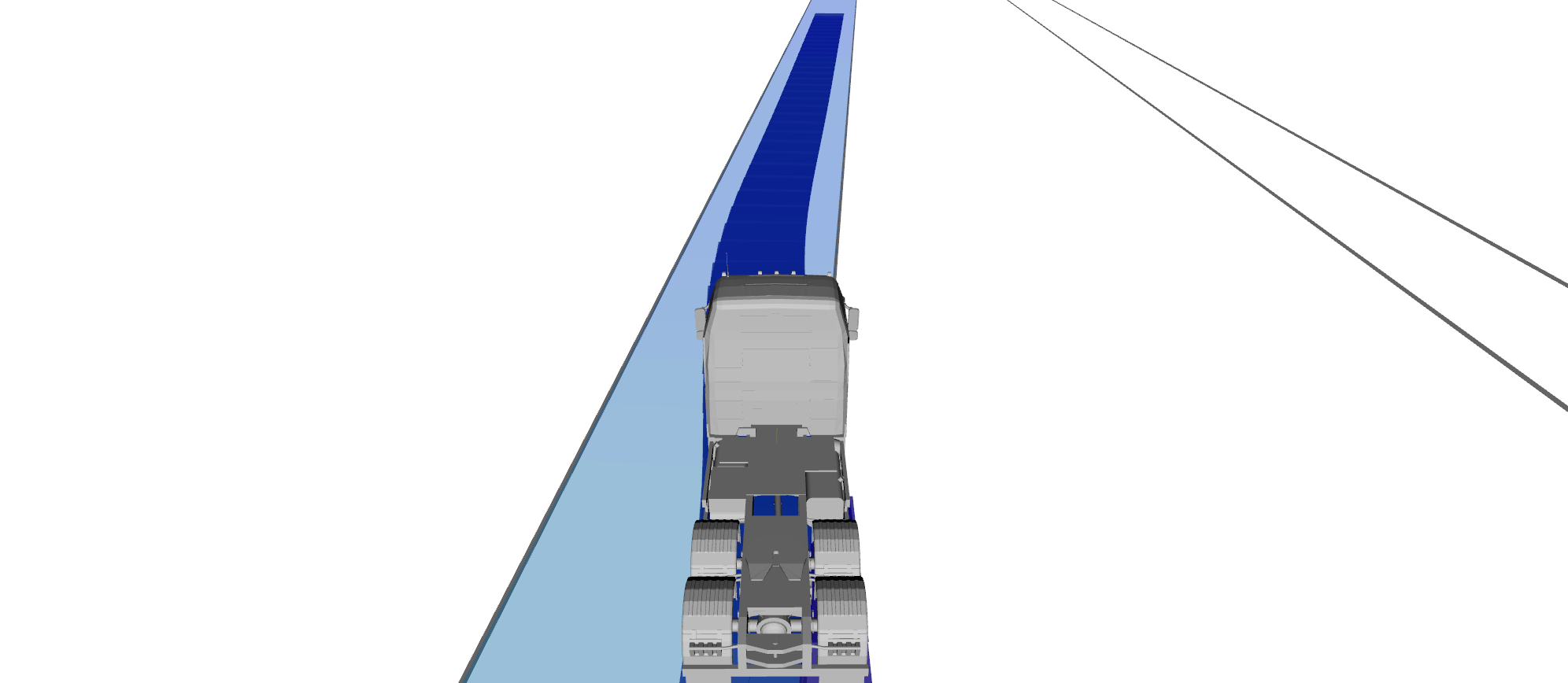}
        \label{fig:collavoid:gp_wce_3}
    }\\
    \subfloat[Distance to obstacle]{%
        \includegraphics[width=0.45\textwidth]{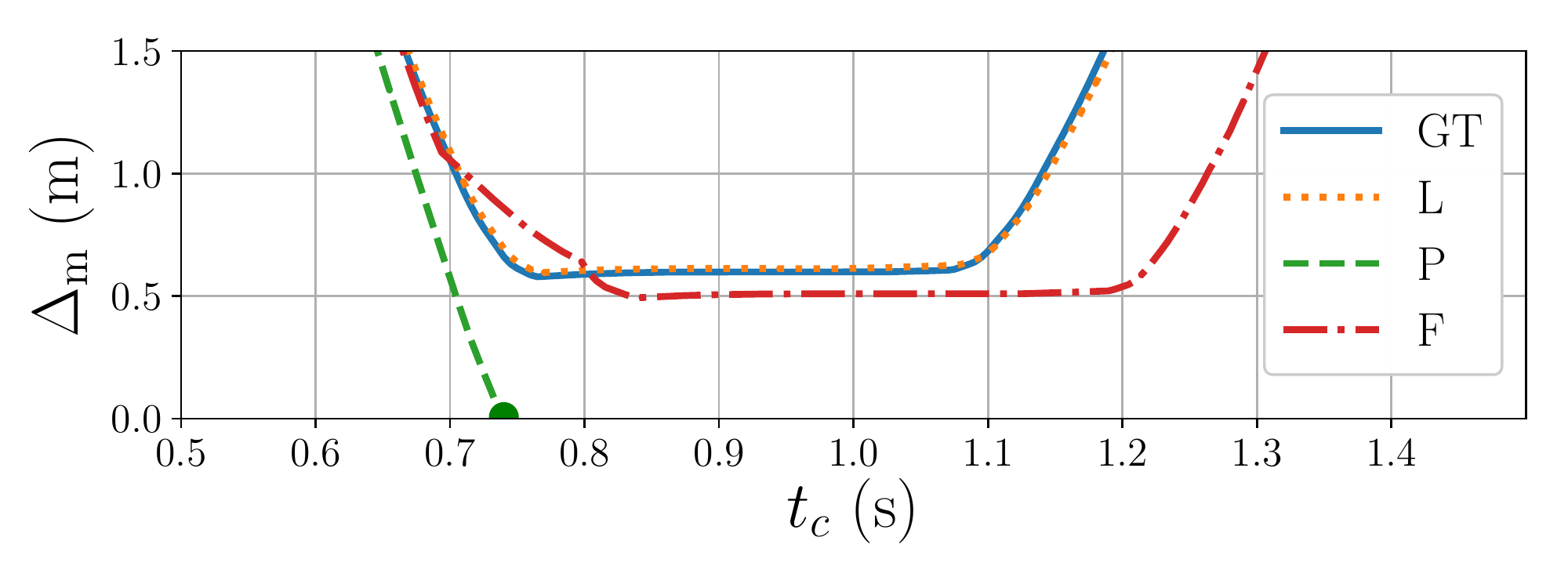}
        \label{fig:collavoid:delta}
    }
    \subfloat[Velocity]{%
        \includegraphics[width=0.45\textwidth]{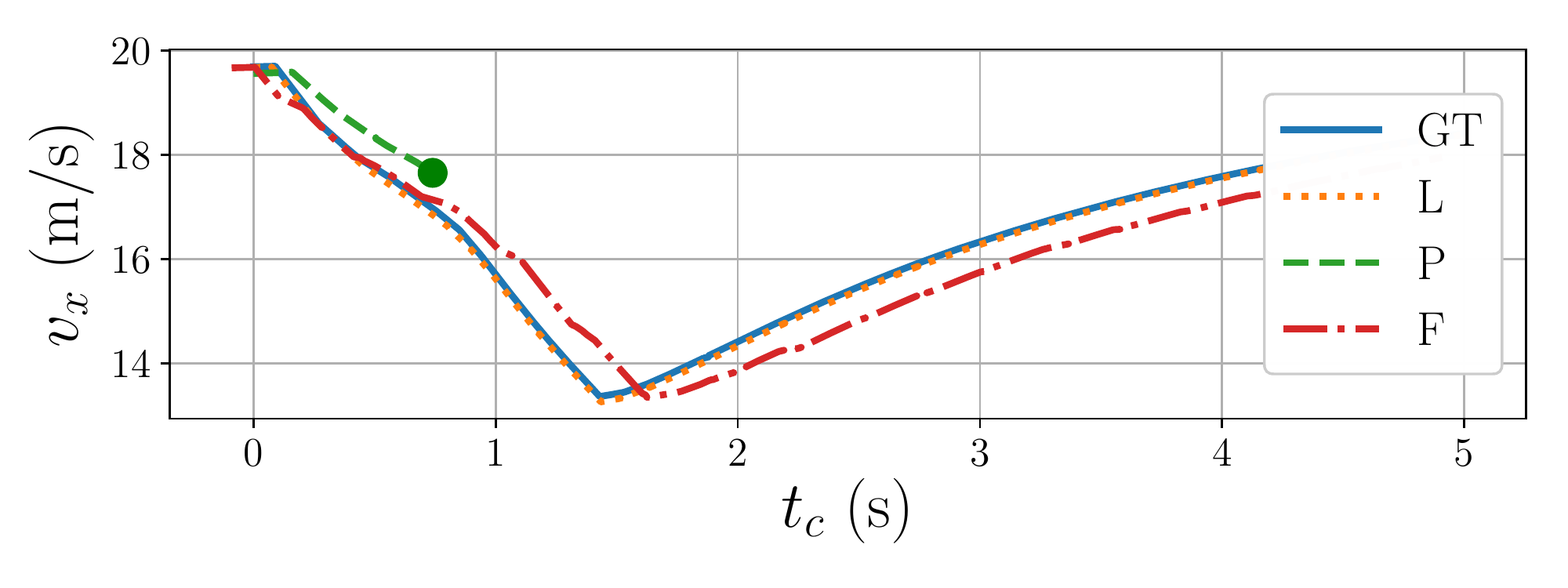}
        \label{fig:collavoid:vx}
    } \\
    \subfloat[Planned front tire forces at $t_c=0.5$]{%
        \includegraphics[width=0.45\textwidth]{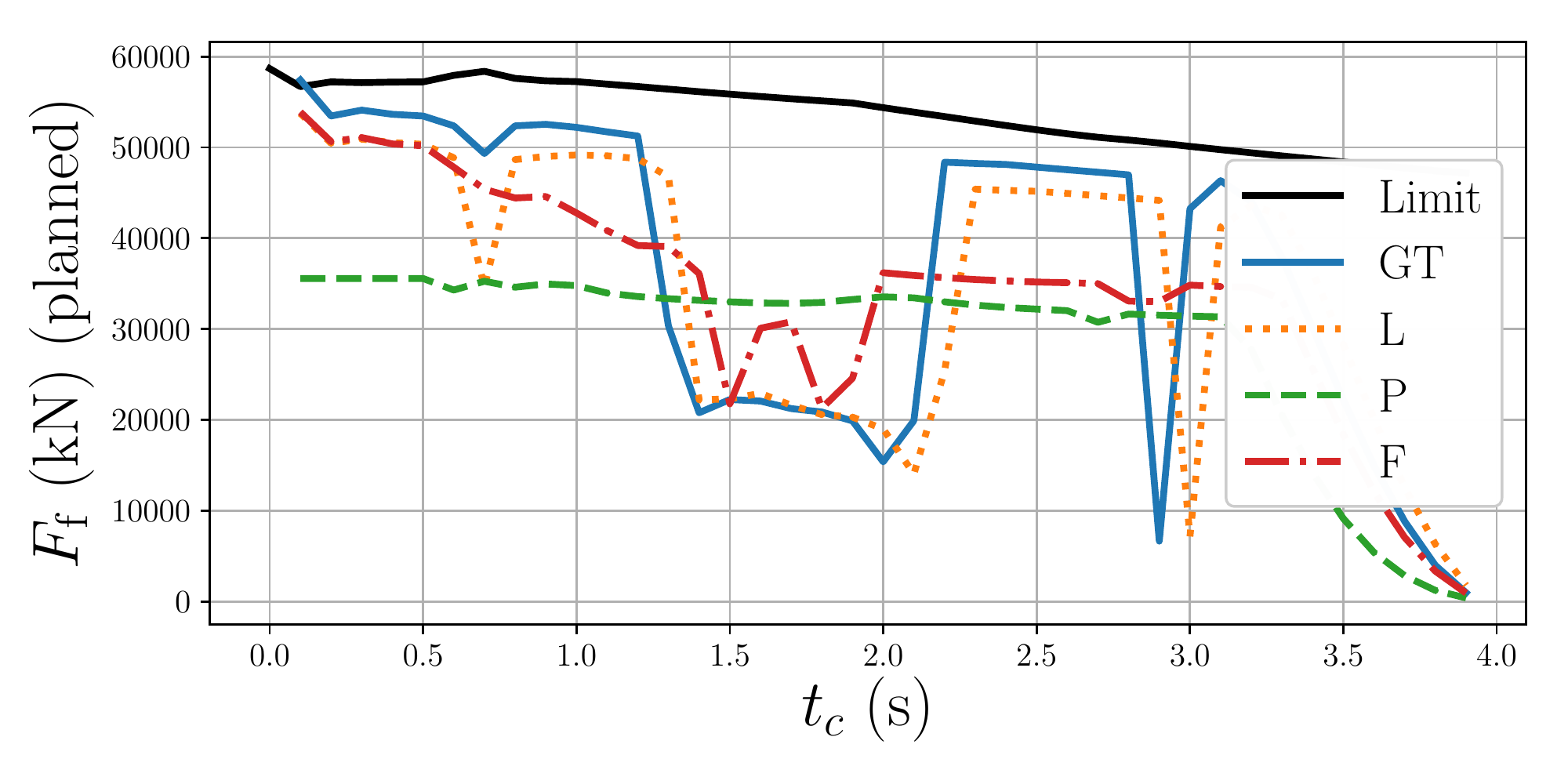}
        \label{fig:collavoid:Ff}
    }
    \subfloat[Merged predictive estimate with uncertainty at $t_c=0.5$]{%
        \includegraphics[width=0.45\textwidth]{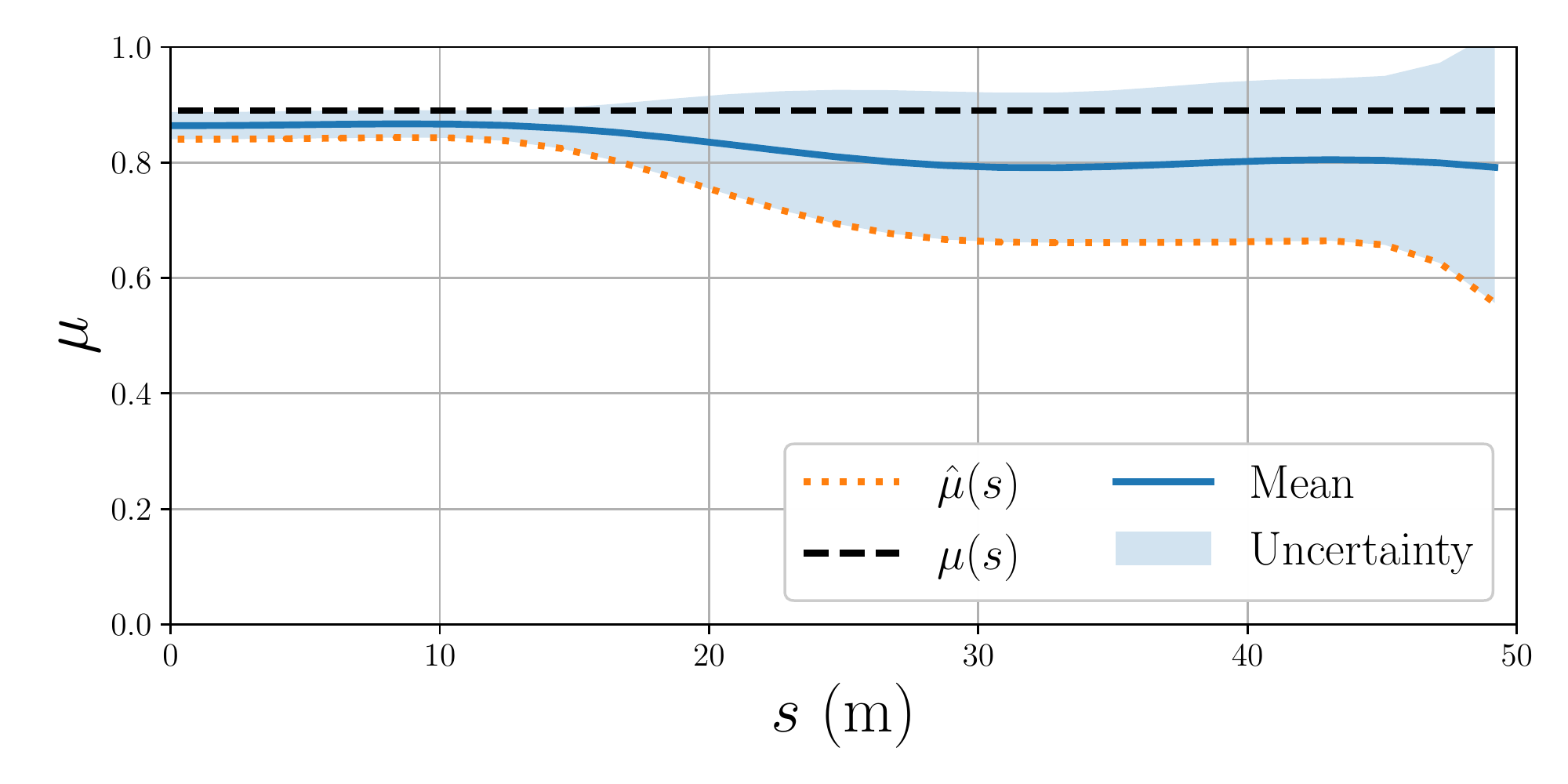}
        \label{fig:collavoid:gp}
    }
    \centering
    \caption{Simulation results for the second critical scenario - collision avoidance at high $\mu$ with worst case under-estimation of friction. Subfigures \ref{fig:collavoid:pred_wce_0} through \ref{fig:collavoid:gp_wce_3} show snapshots from the online visualization of the planned motion. The first row (\ref{fig:collavoid:pred_wce_0} - \ref{fig:collavoid:pred_wce_3}) correspond to the predictive only configuration and the second row (\ref{fig:collavoid:gp_wce_0} - \ref{fig:collavoid:gp_wce_3}) correspond to the proposed fused configuration. 
    Subfigures \ref{fig:collavoid:delta} and \ref{fig:collavoid:vx} show the vehicle's distance to the obstacle and velocity for the four configurations. 
    Subfigure \ref{fig:collavoid:Ff} show planned front tire forces at $t_c=0$s, $s=0$m and the prevailing force limits.
    Subfigure \ref{fig:collavoid:gp} shows the corresponding friction estimate for the fused configuration, including uncertainty.}
    \label{fig:collavoid:full}
\end{figure*}

At the start of the second scenario, $t_c=0$, $s=0$, the vehicle is cruising at a speed of $20$m/s, when an obstacle appears suddenly in the middle of the lane $20$m ahead of the vehicle. The local traction on the road surface is high, i.e., $\mu(s) = 1.0$ for $s \geq 0$m.
Fig.~\ref{fig:collavoid:full}, shows an analysis of the generated motion behavior for the four friction estimation configurations.

% point out that P is different, others are same
Similar to the result of Section \ref{sec:results:scenario_1}, three configurations yield similar behavior and one shows a distinct difference in Figures \ref{fig:collavoid:delta}, \ref{fig:collavoid:vx} and \ref{fig:collavoid:Ff}. However, contrary to the turn scenario, here it is the behavior for configuration P that is distinctly different, while GT, L and F generate similar behavior.
% describe P
Upon detecting the obstacle under the P configuration, Fig.~\ref{fig:collavoid:pred_wce_1}, the vehicle plans an evasive maneuver. However, under the conservative tire force constraints associated with the friction estimate, green dashed line in Fig.~\ref{fig:collavoid:Ff}, no collision free trajectory exists. As such, it selects the least violating option and subsequently collides with the obstacle, Fig.~\ref{fig:collavoid:pred_wce_2}, at an impact velocity of $\sim 17.5$m/s. 

% explain why
The reason behind this undesirable behavior is the overly conservative tire force constraints, which in turn stem from the low accuracy of the predictive friction estimate. In order to ensure that tire forces are never over-estimated, the P configuration selects the lowest friction coefficient value of the associated road surface class. Here, in case of the worst case under-estimation, this corresponds to a $32.6$\% reduction in utilization of available tire force, compared to GT.  

% describe other three
In contrast to P, configurations GT, L and F plans a more aggressive evasive maneuver, as shown in Fig.~\ref{fig:collavoid:delta} and represented by F in Fig.~\ref{fig:collavoid:gp_wce_1}. The vehicle is able to track the more aggressive plan, Fig.~\ref{fig:collavoid:gp_wce_2}, clears the obstacle with a margin of $\sim 0.5$m and safely recovers to the lane, Fig.~\ref{fig:collavoid:gp_wce_3}.   

% explain how accuracy of local fixes the problem
The reason that the L and F configurations perform better than P here is that they benefit from the high accuracy of the local friction estimate. As soon as the vehicle engages in the evasive maneuver, the tire force utilization exceeds the availability threshold, $\lambda > 0.5$ such that the local estimate can be utilized in the next planning iteration.
Since $\max\{|e_l|\} << \max\{|e_p|\}$, the resulting tire force constraints are substantially less conservative ($5.6$\% reduction in utilization of available tire force, compared to GT). 

% comment on GP merge
The L configuration propagates the local estimate and its worst case error over the whole prediction horizon. This turns out to be a good strategy for scenario 2 but a poor one for scenario 1. The F configuration in contrast performs well in both scenarios. Fig.~\ref{fig:lowmuturn:gp} shows the fused friction estimate at $s=0$. The accurate local estimate influences the initial part of the prediction horizon whereas the more conservative predictive estimate dominates the farther part. This way, the fusion strategy exploits the accuracy of the local estimate, while maintaining the foresight of the predictive estimate. 
% comment on delay?

% Discussion on external validity, generalization
\subsection{Discussion}
% generalize from worst case error
The two scenarios selected for evaluation in this paper represent worst case situations with respect to friction estimation error. Hence performance in terms of accident avoidance will not be worse for any estimation error in the intervals $[-\max\{|e_l|\},\max\{|e_l|\}]$, $[-\max\{|e_p|\},\max\{|e_p|\}]$ provided the underlying assumptions on the algorithms, described in Section \ref{sec:results:setup}, hold. 
In scenario 1, in cases where the estimation errors are below the worst case over-estimation, P and F will yield more conservative planned forces, resulting in a larger distance to the traction limit. In scenario 2, in cases where the estimation errors are above the worst case under-estimation, L and F will yield even less conservative planned tire forces, resulting in behavior even closer to that of GT. 

% comment on tuning variables? length scale and s_l
%The method includes two tuning variables that dictate how data  

% comment on compute time? 
%Simulations and the full planning and control stack was run in real-time on regular PC hardware.

% comment on applications?

% MT: The friction estimation function you propose - would be very relevant for lower levels of automation too - would it not? It could for example be used in enhanced ADAS systems. 
% Having good friction estimates can be used for better risk management in general i think, both in terms of proactive safety (tactical safety) as well as for active safety, relevant for assisting /controlling at several of automation levels

\section{Conclusions and Future Work}
\label{sec:conc}

In this work we have highlighted the impact of accuracy, availability and foresight of predictive friction estimation functionality for traction adaptive motion planning in critical situations. Initial simulations with emulated state of the art friction estimation algorithms indicate that fusing heterogeneous friction estimates enables exploiting the virtues of predictive (foresight, high availability) and local (high accuracy) friction estimates. 
Further, we show that the proposed fusion strategy is conservative in that it avoids over-estimation of the friction coefficient while not being overly conservative, even at worst case adversarial estimation errors (e.g., maximum $5.6$\% reduction in tire force utilization, compared to GT, in the collision avoidance scenario). 

% future work
Next steps of this research work is to move development of the method and simulation based evaluation to real world validation on a test vehicle, with state of the art local and predictive friction estimation. Longer term future directions include extending the fusion method to include shared uncertain friction estimates communicated via vehicle-to-vehicle and/or vehicle-to-infrastructure networks.

\section*{Acknowledgements}
This work was partially supported by ECSEL PRYSTINE project, grant agreement 783190.

\bibliographystyle{ieeetr}
\bibliography{refs}

\begin{appendices}

\section{Gaussian Process Regression}
\label{sec:gp_preliminaries}
A Gaussian Process (GP) is a class of random processes, in which any finite collection of random variables have a multivariate normal distribution.  
A GP 
\begin{equation}
    f(x) \sim \mathcal{GP}(\eta(x),k(x,x'))
    \label{eq:gp_prior}
\end{equation}  
is fully specified by its mean function $\eta(x)$ and kernel function $k(x,x')$, the latter specifying the covariance between pairs of random variables.
% Explain GP regression and mention diff to normal regression
Since the distribution of a GP is the joint distribution of possibly infinitely many random variables, it can be interpreted as a distribution over continuous functions. GP regression is the process of taking an existing such distribution over functions (a GP prior) and update it with respect to new data to obtain a refined distribution over functions (a GP posterior). Thus, in contrast to the majority of regression methods, the aim of GP regression is not to fit a single function to data, but rather to fit a distribution of functions to data. 

% The prior
A GP prior is specified by selecting $\eta(x)$ and $k(x,x')$ in \eqref{eq:gp_prior}, based on prior knowledge about the properties of the function we wish to identify. There exists a multitude of alternative kernel functions \cite{rasmussen2003gaussian}. The squared exponential, or radial basis function kernel
\begin{align}
    k(x_i,x_j) = \sigma^2_f \text{exp} \left( -\frac{1}{2l^2}(x_i-x_j)^\top(x_i-x_j) \right),
    \label{eq:squared_exp_kernel}
\end{align} 
used in this project, is one example. Parameters $\sigma_f$ and $l$ are used to tune the smoothness and vertical variation of the prior.  

% posterior computation
The posterior is obtained by conditioning the prior on input data.
Consider $n$ input data points $y(x)$ with uncertainty $\sigma_y(x)$, 
and $n_\star$ test points $x_\star$ at which we wish to predict the mean and covariance of the posterior $f_\star(x_\star)$. The joint distribution of input data and the predicted function values at $x_\star$ is given by 
\begin{equation*}
    \begin{bmatrix}
        y(x) \\
        f_\star(x_\star)
    \end{bmatrix}
    \sim
    \mathcal{N}\left(0,
    \begin{bmatrix}
        K(x,x) + \sigma_y(x) I, ~~K(x,x_\star) \\
        K(x_\star,x), \quad \quad \quad K(x_\star,x_\star) 
    \end{bmatrix}
    \right),
\end{equation*}
where the matrices $K(\cdot,\cdot)$ represent the covariance of all permutations of the input and evaluation points. Applying standard rules for the conditioning of Gaussians \cite{rasmussen2003gaussian} yields the posterior predictive distribution 
\begin{align}
    &f_\star(x_\star) \sim \mathcal{GP}(\eta_\star(x_\star),\Sigma_\star) \quad \text{with} \\
    &\eta_\star(x_\star) = K(x_\star,x) (K(x,x) + \sigma_y(x)^2 I)^{-1} y(x), \label{eq:gp_post_mean} \\
    &\Sigma_\star = K(x_\star,x_\star) - K(x_\star,x) (K(x,x) + \sigma_y(x)^2 I)^{-1} K(x,x_\star), \label{eq:gp_post_cov}
\end{align}
where $\Sigma_\star$ denotes the covariance matrix of the posterior distribution.

% heteroscedasticity
Standard GP regression builds on the assumption that all measurements have the same measurement noise, i.e., that $\sigma_y(x)=\sigma_y$ is constant for all $x$. This limitation is lifted by Heteroscedastic GP regression. The extension is described e.g., in \cite{goldberg1998regression,lazaro2011variational}.

\end{appendices}

\end{document}